\documentclass[10pt,twocolumn,letterpaper]{article}

\usepackage[pagenumbers]{cvpr} %

\providecommand{\color}[1]{}
\usepackage{xcolor}

\usepackage{pdflscape}

\usepackage[utf8]{inputenc}
\usepackage{tcolorbox}
\tcbuselibrary{listings,skins,breakable}

\usepackage{makecell}
\usepackage{float}

\usepackage{soul}

\newtcblisting{promptbox}{
  breakable,
  colback=black!2,
  colframe=black!20,
  arc=10pt,
  boxrule=0.8pt,
  left=5pt,
  right=5pt,
  top=3pt,
  bottom=0pt,
  listing only,
  listing engine=listings, %
  listing options={
    basicstyle=\ttfamily\small,
    breaklines=true,
    showstringspaces=false,
   escapeinside={(*@}{@*)} %
  }
}

\definecolor{cvprblue}{rgb}{0.21,0.49,0.74}
\usepackage[pagebackref,breaklinks,colorlinks,allcolors=cvprblue]{hyperref}

\def\paperID{39888} %
\def\confName{CVPR}
\def\confYear{2026}

\title{
Counterfactual VLA: Self-Reflective Vision-Language-Action Model\\
with Adaptive Reasoning
}

\begin{small}
\author{
Zhenghao ``Mark'' Peng\textsuperscript{1,2}\thanks{Work done while the author was an intern at NVIDIA.} \quad
Wenhao Ding\textsuperscript{1} \quad
Yurong You\textsuperscript{1} \quad
Yuxiao Chen\textsuperscript{1} \quad
Wenjie Luo\textsuperscript{1}
\\
Thomas Tian\textsuperscript{1} \quad
Yulong Cao\textsuperscript{1}  \quad
Apoorva Sharma\textsuperscript{1} \quad
Danfei Xu\textsuperscript{1} \quad
Boris Ivanovic\textsuperscript{1} \quad
\\
Boyi Li\textsuperscript{1} \quad
Bolei Zhou\textsuperscript{2} \quad
Yan Wang\textsuperscript{1}\thanks{Equal advising.} \quad
Marco Pavone\textsuperscript{1,3}\footnotemark[2] \\
\normalsize{
\textsuperscript{1}NVIDIA \qquad
\textsuperscript{2}UCLA \qquad
\textsuperscript{3}Stanford University}
}
\end{small}

\begin{document}
\maketitle

\begin{abstract}

Recent reasoning-augmented Vision-Language-Action (VLA) models have improved the interpretability of end-to-end autonomous driving by generating intermediate reasoning traces. Yet these models primarily describe what they perceive and intend to do, rarely questioning whether their planned actions are safe or appropriate. This work introduces Counterfactual VLA (CF-VLA), a self-reflective VLA framework that enables the model to reason about and revise its planned actions before execution. CF-VLA first generates time-segmented meta-actions that summarize driving intent, and then performs counterfactual reasoning conditioned on both the meta-actions and the visual context. This step simulates potential outcomes, identifies unsafe behaviors, and outputs corrected meta-actions that guide the final trajectory generation. To efficiently obtain such self-reflective capabilities, we propose a rollout–filter–label pipeline that mines high-value scenes from a base (non-counterfactual) VLA's rollouts and labels counterfactual reasoning traces for subsequent training rounds. Experiments on large-scale driving datasets show that CF-VLA improves trajectory accuracy by up to 17.6\%, enhances safety metrics by 20.5\%, and exhibits adaptive thinking: it only enables counterfactual reasoning in challenging scenarios. By transforming reasoning traces from one-shot descriptions to causal self-correction signals, CF-VLA takes a step toward self-reflective autonomous driving agents that learn to think before they act.

\end{abstract}

\section{Introduction}

Recent advances in Vision–Language–Action (VLA) models highlight the promise of test-time reasoning for embodied decision-making. 
By generating intermediate language traces that describe the scene and task, reasoning-augmented VLAs have improved interpretability and robustness in manipulation~\cite{team2025gemini,bjorck2025gr00t,chen2025internvla,intelligence2504pi0} as well as autonomous driving~\citep{nvidia2025alpamayo,luo2025adathinkdrive,renz2025simlingo}. 
In these systems, a large vision-language backbone engages in a slower, more deliberative form of ``thinking'', spending additional compute to verbalize the observation and justify the planned actions.

\begin{figure}[!t]
    \centering
    \vspace{-4mm}
    \includegraphics[width=\linewidth]{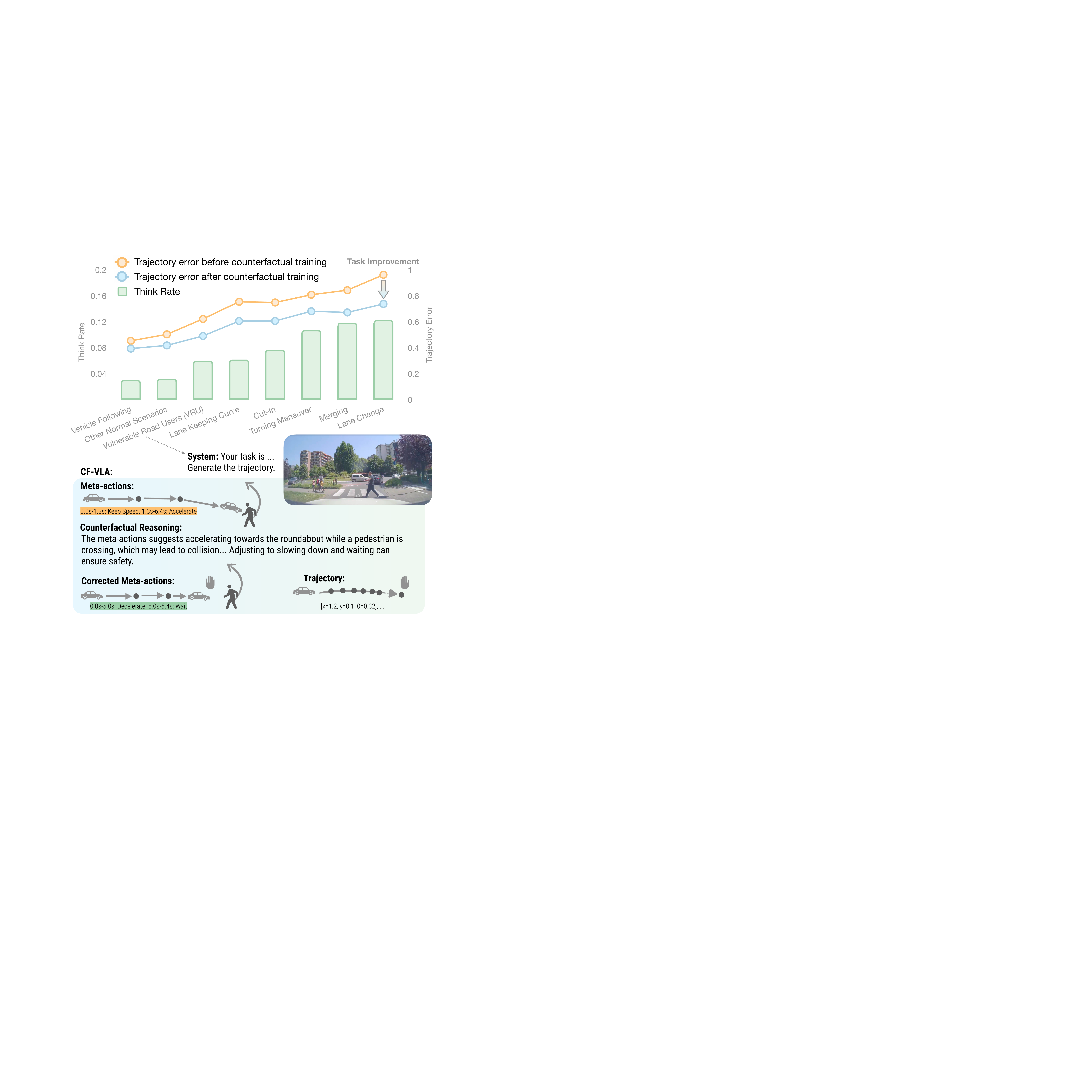}
    \vspace{-6mm}
    \caption{
    \textbf{Counterfactual Vision-Language-Action (CF-VLA) Model.}
    Top: CF-VLA conducts reasoning adaptively. 
    The model engages in reasoning more frequently and achieves more significant task performance gains in complex scenarios that have higher trajectory errors.
    Bottom: CF-VLA reflects on its own action plan and corrects it before generating the final trajectory.
    }
    \label{fig:teaser}
    \vspace{-10pt}
\end{figure}

However, the reasoning in current VLAs is largely {descriptive} rather than \emph{\textbf{self-reflective}}. Existing models typically describe what they observe (e.g., ``the cabbage next to the bowl'' and ``a pedestrian is crossing'') and what they intend to do (e.g., ``place the cabbage into the container'' and ``I should be cautious''). As a result, the reasoning trace often serves as a one-shot commentary on the scene and action choice, without a self-reflective loop that verifies whether the model’s own command is appropriate: once the VLA model produces a textual intent, it is typically treated as ground truth and used to condition a low-level policy, rather than being checked for inconsistencies with visual cues and revised accordingly.

Self-correction has been explored in embodied VLMs through replanning and failure recovery~\citep{team2025gemini,huang2025thinkact},
where the agent detects that an action it executed failed and then switches to an alternative plan. 
These mechanisms typically operate \emph{after} observing a mistake or via an external verifier, and do not enable the VLA itself to reason explicitly about the consequences of its own action plan \emph{before} execution. 
We refer to this desired ability as \textbf{\textit{counterfactual reasoning}}.
More recent robotics VLAs begin to incorporate world models to proactively simulate, verify, and select planning steps~\cite{wu2025foresight,qi2025strengthening,chen2025reimagination}. While these approaches move toward earlier detection of problematic actions, they fundamentally rely on an external future-prediction model to judge the quality of the proposed plan. This is qualitatively different from self-reflection: external simulation can evaluate a plan, but it cannot help the VLA understand its own reasoning process.
This raises a central question: \textit{can we obtain self-reflective, counterfactual reasoning {inside} the VLA itself, without an external world model or verifier, analogous to the self-reflection behaviors observed in language-only reasoning models?}
Two obstacles make this challenging.
First, most VLAs lack an \textit{action--language alignment}: the action is represented by latent tokens~\cite{intelligence2025pi_}, and there is no action$\rightarrow$language alignment, leaving the language model with no handle to talk about its actions. 
Second, standard training pipelines rarely teach models to answer \textit{counterfactual questions} such as: {``Given the plan I just proposed, what will happen, and how should I change it?''}

In this paper, we propose \textbf{\textit{Counterfactual VLA (CF-VLA)}}, a VLA equipped with a \emph{self-reflective} reasoning loop that performs counterfactual analysis directly on its predicted controls. 
As illustrated in \cref{fig:teaser}~(bottom), CF-VLA first predicts a sequence of language-based, time-segmented meta-actions that summarize the agent’s intent.
Instead of treating them as final, the model then conditions a counterfactual chain-of-thought step on both the visual context and its own meta-actions, asking ``If I follow this plan, what would happen and is that desirable?'' 
It revises unsafe or suboptimal plans (e.g., from ``accelerate toward the intersection'' to ``decelerate early and yield'') before committing to the final trajectory. 
This meta-actions $\rightarrow$ counterfactual reasoning $\rightarrow$ updated meta-actions $\rightarrow$ trajectory loop upgrades reasoning from one-shot description to counterfactual analysis of the model’s own behavior, and turns that one-shot analysis into actionable self-correction.
To realize such behavior in practice, CF-VLA combines meta-actions with a rollout–filter–label pipeline:
1) the current policy is rolled out to generate candidate meta-actions and trajectories; 
2) high-value data points are automatically selected by checking if pre-filling ground-truth meta-actions substantially improves trajectories over model-generated meta-actions; and
3) a teacher model is prompted to produce counterfactual reasoning traces that explain why the current plan is less preferable and how to adjust it. 
Training on a mixture of nominal and counterfactual-labeled datasets under a unified instruction prompt yields a single CF-VLA model that exhibits \textbf{\textit{adaptive counterfactual reasoning}}: 
as shown in \cref{fig:teaser} (top), the model thinks more often and achieves larger task improvement in the hardest scenarios.

We conduct extensive experiments on a large-scale internal dataset to validate our design. CF-VLA consistently outperforms trajectory-only and non-reflective meta-action baselines on trajectory metrics by 17.6\% and 9\%, respectively, and increases the safety metric by 14.7\%.
We further show that performance can be improved if we apply the trained CF-VLA again to the rollout-filter-label pipeline for multi-round training.
Our core contributions are:
\begin{enumerate}[label=\textbf{\arabic*)}, leftmargin=8pt, itemsep=3pt, topsep=1pt]
    \item \textbf{Self-reflective counterfactual reasoning for VLA.} 
    We propose a new reasoning-about-action paradigm to condition the reasoning of VLA on its own predicted meta-actions, anticipate the consequences, and revise the plan before generating final actions. This upgrades reasoning from descriptive explanations to causal self-correction.
    \item \textbf{Meta-actions and counterfactual data pipeline.} 
    We use time-segmented meta-actions for action-language alignment and propose the rollout–filter–label pipeline to automatically curate counterfactual data from the rollout of the model, forming a self-improving loop that enhances both reasoning and action.
    \item \textbf{Adaptive thinking in autonomous driving.} 
    CF-VLA exhibits the ability to \emph{think when necessary}, concentrating counterfactual reasoning on the most challenging scenarios. Experiments show that CF-VLA improves trajectory accuracy, safety metrics, and meta-action alignment, while maintaining reasonable test-time compute by adapting its think rate to the scenario difficulty.
\end{enumerate}

\section{Related Works}

\noindent \textbf{VLA Models for Autonomous Driving.}
Recent VLA models integrate language reasoning with visual context to generate control actions. Some use high-level commands~\cite{jiang2025alphadrive,xu2024drivegpt4,shao2024lmdrive}, while others rely on intermediate trajectory tokens~\cite{tian2025drivevlm,zhou2025opendrivevla,renz2025simlingo,nvidia2025alpamayo}. SimLingo~\cite{renz2025simlingo} introduces ``action dreaming'' to align language and control via offline counterfactual simulation. Alpamayo-R1~\cite{nvidia2025alpamayo} incorporates a structured language abstraction and trajectory diffusion to improve long-tail generalization. 
AutoVLA~\cite{zhou2025autovla} incorporates chain-of-thought (CoT) traces, but uses them primarily as one-pass rationales for interpretability.
These models enhance interpretability but remain largely descriptive: they describe what the model sees and intends, without critiquing the proposed action itself.  In contrast, our CF-VLA explicitly conditions reasoning on the model’s own proposed behavior, enabling self-reflective revision through counterfactual analysis.

\noindent \textbf{Counterfactual Reasoning and Self-Reflection.}
Counterfactual reasoning models alternative actions and their consequences, but most prior work uses counterfactuals externally rather than as model-internal self-reflection. CAST~\cite{glossop2025cast}, SimLingo~\cite{renz2025simlingo}, and VLAPS~\cite{neary2025improving} generate counterfactual instruction–action pairs or search over simulated futures to improve instruction following or correct policies, while the underlying VLA does not itself reason about or revise its own plan. A separate line of work introduces reflective loops around the policy using auxiliary verifiers or planners~\cite{huang2025thinkact,wu2025foresight,chen2025reimagination}. 
Several works analyze failures and successes with large models to relabel data, synthesize rewards, or trigger replanning~\cite{liu2023reflect,abdolmaleki2025gemini,ghasemipour2025self,li2025reflection}.
In all these cases, counterfactual evaluation and reflection are implemented by external modules wrapped around the policy. In contrast, CF-VLA performs counterfactual reasoning over its \emph{own} predicted meta-actions inside the forward pass and revises them before trajectory generation, yielding an online self-correction mechanism without separate validators.

\begin{figure}[!t]
    \centering
    \includegraphics[width=1.0\linewidth]{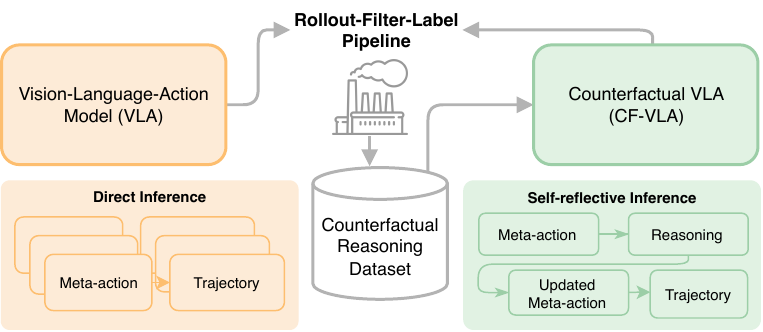}
    \vspace{-6mm}
    \caption{\textbf{The framework of CF-VLA.} 
A base VLA is fine-tuned on a counterfactual reasoning dataset generated by a rollout–filter–label pipeline. The resulting CF-VLA supports both direct inference and self-reflective inference, in which counterfactual reasoning edits meta-actions before trajectory generation.
    }
    \label{fig:framework}
    \vspace{-10pt}
\end{figure}

\noindent \textbf{Adaptive Reasoning.} 
Adaptive reasoning dynamically selects between direct response and step-by-step thinking~\cite{zhang2025adaptthink,wu2025arm,wang2025adareasoner,zhan2025kat,yu2025think,chen2024toward,chung2025thinker}. 
OneTwoVLA~\citep{lin2025onetwovla} uses control tokens to toggle between fast acting and slow reasoning, invoking language reasoning mainly at subtask boundaries (e.g., task completion or errors), so “slow thinking’’ is tied to task switching rather than scene difficulty. 
In contrast, CF-VLA learns to invoke reasoning from visual cues and its own behavior, focusing on challenging scenes and skipping reasoning in normal cases. 
AdaThinkDrive~\citep{luo2025adathinkdrive} categorizes scenarios into simple or challenging ones via rule-based heuristics and uses RL to fine-tune when to think. 
We instead show that adaptive reasoning can emerge from supervised fine-tuning alone: by identifying hard scenarios based on the model’s capability (\cref{sec:counterfactual}) and training on a mixture of data, CF-VLA naturally allocates more test-time compute to difficult scenes (see~\cref{fig:teaser}).

\section{Method}

End-to-end Vision-Language-Action (VLA) models have demonstrated promising progress in mapping visual context directly to control output. However, their reasoning processes are largely descriptive: when the agent proposes an incorrect plan, there is no mechanism for the model itself to analyze the plan and revise the decision before execution. 
This section introduces \textbf{\textit{Counterfactual VLA (CF-VLA)}}, which equips a VLA with a self-reflective loop that reasons about its own predicted actions and uses that reasoning to correct the plan.

\begin{figure*}[!t]
    \centering
    \includegraphics[width=1.0\linewidth]{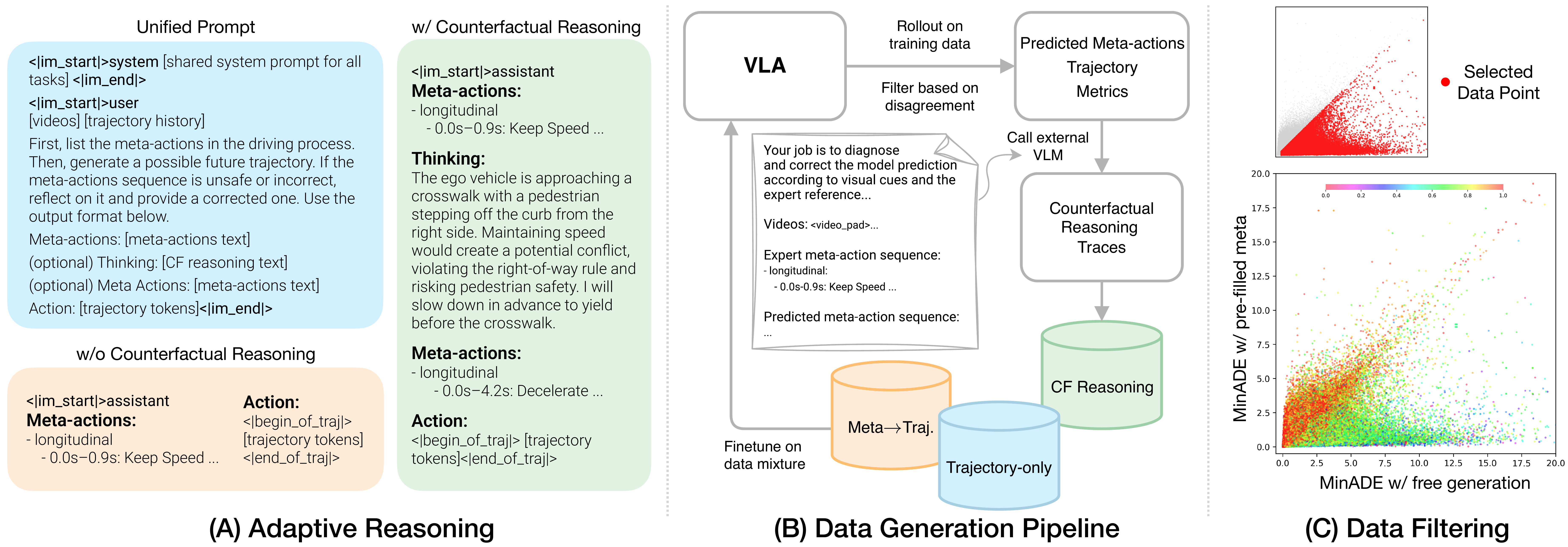}
    \vspace{-1.5em}
    \caption{
        \textbf{(A) Adaptive Reasoning} can be achieved by training models on a mixture of data with the unified instruction prompt.
    \textbf{(B) Data generation process.} 
    We build a rollout–filter–label pipeline that runs the VLA, detects samples where its meta-actions are problematic, and labels counterfactual (CF) reasoning traces, forming a CF reasoning dataset. 
    \textbf{(C) Data filtering process.} 
    We use {trajectory disagreement} between trajectories that are free-generated and those induced by the ground-truth meta-actions to filter data.
    Each data point is colored by the meta-actions IOU in free generation.
    }
    \label{fig:data-pipeline}
    \vspace{-10pt}
\end{figure*}

\subsection{Self-Reflective Counterfactual Reasoning}
\label{sec:counterfactual}

Several challenges remain to empower a VLA with counterfactual (CF) reasoning. 
First, the model needs an intermediate representation that is both interpretable to the language backbone and tightly coupled to action. We address this with time-segmented meta-actions (\cref{sec:meta-actions}), so the model can reason about and revise high-level intent in language space before decoding trajectories. 
Second, CF reasoning must relate meta-actions to their future consequences. 
To prepare data to fine-tune the model, we develop a {rollout--filter--label} pipeline (\cref{sec:rollout-filter-label}) that selects data and auto-generates high-value counterfactual reasoning traces.
We consider CF reasoning as a {plugged-in self-reflective mechanism} on top of meta-actions. As shown in \cref{fig:framework}, instead of mapping meta-actions to trajectories (\texttt{meta$\rightarrow$traj}), CF-VLA performs a self-reflective loop:
\begin{center}
\vspace{-5pt}
\texttt{meta-actions}
$\rightarrow$
\texttt{CF reasoning}\\
$\rightarrow$
\texttt{updated meta-actions}
$\rightarrow$
\texttt{trajectory}.
\vspace{-2pt}
\end{center}

\noindent \textbf{Adaptive Thinking.} 
Adaptive reasoning allows a model to decide dynamically when to conduct reasoning and when to respond directly. This mechanism is essential because most scenarios are straightforward, and explicit reasoning on them increases hallucination risks and wastes test-time compute.
As shown in \cref{fig:data-pipeline}(A), we use the same instruction for the model and allow it to implicitly decide whether to generate the reasoning traces. 
As both meta-actions and reasoning operate in the language space, the CF reasoning behavior is controlled by the words (\texttt{Action:} or \texttt{Thinking:}) generated after the first meta-action sequence.
Training on a mixture of samples with and without CF traces allows the model to implicitly learn when self-reflective inference is necessary.

\subsection{Meta-Actions}
\label{sec:meta-actions}

Meta-actions provide a language-native intermediate abstraction between reasoning and low-level action. 
Each meta-action sequence expresses the ego vehicle’s intended behavior along three orthogonal dimensions:
\textit{longitudinal} ({Accelerate, Decelerate, Keep Speed, Wait, Reverse}),
\textit{lateral} ({Straight, Left Turn, Right Turn}), and
\textit{lane-level} ({Keep Lane, Left Lane Change, Right Lane Change}).

Though playing an analogous role to the low-level command used in manipulation VLAs~\cite{team2025gemini,intelligence2025pi_}, navigation models~\cite{cheng2024navila} and autonomous driving VLAs~\cite{tian2025drivevlm,zhou2025autovla}, our meta-actions consider timing and are tightly coupled to the continuous trajectories.
CF-VLA models meta-actions as
\textbf{time-partitioned segments} covering the 6.4-second planning horizon.
Within each of the three groups, meta-actions are defined over non-overlapping temporal intervals
and together describe the intended evolution of driving behavior.
This temporal format allows the model to reason compositionally about action transitions and to capture temporal intent and directly align language reasoning with the structure of the predicted trajectory. 
Example meta-action sequences can be found in \cref{fig:qualitative}.

\subsection{Rollout-Filter-Label Counterfactual Pipeline}
\label{sec:rollout-filter-label}

As illustrated in \cref{fig:data-pipeline}(B), to supervise counterfactual reasoning, CF-VLA relies on a
{rollout--filter--label} data curation pipeline that mines
high-value scenes from the model's own behavior.

\noindent \textbf{Data Rollout.}
Starting from a VLA trained with meta-actions but without
counterfactual reasoning, the model is rolled out on the
training set. For each scene, two sets of trajectories are
generated: 
1) \textit{Free generation} $\mathbf{x}_\text{free}$:
  the model first predicts meta-actions and then decodes
  the trajectory conditioned on its \emph{own} meta-actions.
2) \textit{Pre-filled meta-actions}
  $\mathbf{x}_\text{pf}$:
  the model is conditioned on the \emph{ground-truth} meta-actions
  and only decodes the trajectory.
For robustness, 6 output trajectories are sampled per
scene in each setting. This yields paired sets of trajectories
$(\mathbf{x}_\text{free}, \mathbf{x}_\text{pf})$ for the same
visual context.

\noindent \textbf{Data Filtering.}
Let $\text{minADE}(\mathbf{x},
{x}^\star)$ denote the minimum displacement error
between a set of predicted trajectories $\mathbf{x}$ and the
expert future ${x}^\star$. 
We scatter each scenario based on $(
\text{minADE}(\mathbf{x}_\text{free}, {x}^\star),
\text{minADE}(\mathbf{x}_\text{pf}, {x}^\star)
)$
in \cref{fig:data-pipeline}(C), where each data point is colored by the meta-actions accuracy (IOU) in free generation.
The key insight is that many scenes lie below the diagonal: the model performs poorly in free
generation but matches the expert trajectory when meta-actions are pre-filled and they usually have low IOU.
These are precisely the scenes where meta-actions are the bottleneck. 
CF-VLA uses \textbf{trajectory disagreement} between trajectories that are free-generated and those induced by ground-truth-filled meta-actions to filter data:
$
\text{minADE}(\mathbf{x}_\text{pf}, {x}^\star)
<
\text{minADE}(\mathbf{x}_\text{free}, {x}^\star)
$ and $\text{minADE}(\mathbf{x}_\text{free}, {x}^\star) > \epsilon$, where $\epsilon=0.5$ avoids labels on already mastered scenarios.
Intuitively, these are cases where improving meta-actions would significantly improve trajectory quality.
As long as we improve the meta-actions to be closer to the ground truth, we can harvest task improvement.
Samples above the diagonal already have good free-generation trajectories, so counterfactual supervision brings limited benefit and is not needed. \cref{sec:ablation-studies} shows the data filtering is crucial to the final performance.

\noindent \textbf{Data Labeling.}
For the filtered scenes, concise counterfactual traces are
generated with a high-capacity teacher model
(\texttt{Qwen2.5-VL-72B-Instruct}). 
The outline of the prompt to the teacher is presented in \cref{fig:data-pipeline}(B).
The output is a single paragraph that:
1) diagnoses why the predicted meta-actions are less preferable than the expert plan, and 2) indicates how they should be adjusted.
These labeled samples form the {counterfactual
reasoning dataset} $\mathcal{D}_\text{CF}$. The full teacher prompt
is provided in the Appendix.

\begin{figure}[!t]
    \centering
    \includegraphics[width=0.8\linewidth]{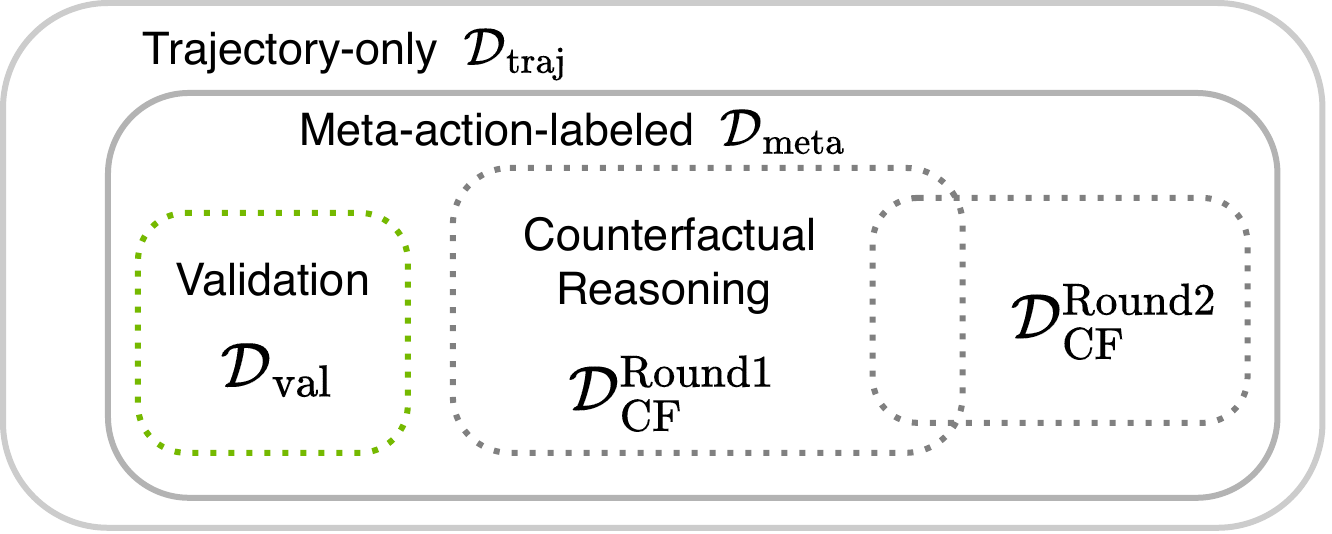}
    \vspace{-0.5em}
    \caption{
    \textbf{ The dataset composition.} We use a subset of the meta-action-labeled dataset $\mathcal{D}_\text{meta}$ as the validation set $\mathcal{D}_\text{val}$.
    }
    \label{fig:data_exp}
    \vspace{-15pt}
\end{figure}

\subsection{Implementation Details}

\noindent \textbf{Mixed-data Training.}
We adopt a mixed-data training scheme combining the trajectory-only dataset $\mathcal{D}_\text{traj}$, the meta-action–labeled dataset $\mathcal{D}_\text{meta}$, and the counterfactual reasoning dataset $\mathcal{D}_\text{CF}$ as shown in \cref{fig:data_exp}. Training proceeds in stages. The base VLM is first trained on $\mathcal{D}_\text{traj}$ to learn basic trajectory generation (\texttt{traj-only}). Meta-actions are then introduced by fine-tuning on $\mathcal{D}_\text{traj} \cup \mathcal{D}_\text{meta}$, yielding
the \texttt{meta-act} model used in the initial rollout
(Sec.~\ref{sec:rollout-filter-label}). Finally, the full CF-VLA is
obtained by further fine-tuning on the mixture
$\mathcal{D}_\text{traj} \cup \mathcal{D}_\text{meta} \cup \mathcal{D}_\text{CF}$.
We unfreeze all parameters during training.

\noindent \textbf{Loss Masking and Weighting.}
The model is optimized with cross-entropy
loss over assistant-generated tokens only; tokens from
system or user prompts are masked. For counterfactual
samples in $\mathcal{D}_\text{CF}$, the loss on the \emph{first}
(uncorrected) meta-action block is also masked to prevent the model from learning from prior mistakes.
Within the assistant response, different token groups (meta-actions, reasoning, or trajectory tokens) have different loss weights.

\noindent \textbf{Multi-round training.}
An appealing property of CF-VLA is that the trained model
can be inserted back into the rollout--filter--label loop to
create new rounds of CF data
$\mathcal{D}_\text{CF}^\text{Round2}$.
Unlike traditional CoT methods that generate mostly
deterministic explanations for a given scene, CF-VLA's
reasoning is conditioned on the predicted
meta-actions and can therefore produce diverse reasoning traces for the same scenario. This allows us to further exploit the dataset and generate different reasoning traces with different meta-actions.
As shown in \cref{sec:main-exp}, fine-tuning on another round of CF datasets further improves the model, realizing a
self-improving counterfactual flywheel.

\noindent \textbf{Model Architecture.}
CF-VLA is similar in size and design to Alpamayo-R1~\cite{nvidia2025alpamayo}.
The model takes text prompts, two front-facing videos, and the ego-trajectory history as inputs.
The text prompt defines the tasks: 1) pure trajectory prediction, or 2) meta-actions and trajectory prediction with optional counterfactual reasoning.
A wide (120°) and a telephoto (30°) cameras provide 2 videos at 2~Hz over the past 2~s.
The past 1.6~s of ego motion is embedded into a single trajectory-history token by an MLP-based {trajectory history encoder}.
Future motion is represented by a compact set of {discrete trajectory tokens}. We expand the vocabulary of the VLM backbone to accommodate the new tokens introduced by the trajectory tokenizer, as well as additional \texttt{$<$begin\_of\_traj>} and \texttt{$<$end\_of\_traj>} tokens.

\section{Experiments}

\begin{table*}[!t]
\centering
\caption{\textbf{Evaluation results.} 
CF-VLA improves trajectory accuracy (ADE, FDE), behavioral safety (Corner Distance, Collision, Off-road), and reasoning quality (IOU).
$\downarrow$ lower is better, $\uparrow$ higher is better.
}
\label{tab:openloop_results}
\vspace{-2mm}
\small
{
\begin{tabular}{lccccccc}
\toprule
\textbf{Model} &
\begin{tabular}[c]{@{}c@{}}\textbf{ADE$\downarrow$} \\Min (Avg)\end{tabular} 
&
\begin{tabular}[c]{@{}c@{}}\textbf{FDE$\downarrow$} \\Min (Avg)\end{tabular} 
&
\begin{tabular}[c]{@{}c@{}}\textbf{Corner}\\\textbf{Dist.$\downarrow$}\end{tabular} &
\textbf{Collision$\downarrow$} &
\textbf{Off-road$\downarrow$} &
\begin{tabular}[c]{@{}c@{}}\textbf{IOU$\uparrow$}  \\init$\rightarrow$edited\end{tabular} 
&
\begin{tabular}[c]{@{}c@{}}\textbf{Output Len.}\\\textbf{(Think Rate)}\end{tabular}
\\
\midrule

\scriptsize{\texttt{traj-only}} &
0.9283 (1.8284) &
2.5912 (5.1150) &
0.8563 &
0.0244 &
0.0720
&
-- &
10.00 (--) \\

\midrule

\scriptsize{\texttt{lang-meta-act}} &
0.8021 (1.5607) &
2.2540 (4.4967) &
0.7358 &
0.0206 &
0.0617 &
0.9183 &
144.28 (1.00) \\

\midrule

\scriptsize{\texttt{meta-act} (w/o route)} &
0.8411 (1.6216) &
2.3647 (4.6616) &
0.7720 &
0.0224 &
0.0625 &
0.9169 &
85.32 (--) \\

\scriptsize{\texttt{CF-VLA} (w/o r., round1)} &
0.7650 (1.5606) &
2.1416 (4.3307) &
\textbf{0.6975} &
\textbf{0.0191} &
0.0601 &
0.9153$\rightarrow$0.9212 &
113.36 (0.148) \\

\scriptsize{\texttt{CF-VLA} (w/o r., round2)} &
\textbf{0.7647} (\textbf{1.5032}) &
\textbf{2.1365} (\textbf{4.1927}) &
0.6996 &
0.0194 &
\textbf{0.0583} &
\textbf{0.9174}$\rightarrow$\textbf{0.9228} &
102.12 (0.083) \\

\midrule

\scriptsize{\texttt{meta-act} (w/ route)} &
0.7263 (1.4612) &
1.9561 (3.9269) &
0.6600 &
0.0196 &
0.0619 
&
0.9236 &
87.20 (--) \\

\scriptsize{\texttt{CF-VLA} (w/ r., round1)} &
\textbf{0.6712} (1.4574) &
\textbf{1.7988} (3.9466) &
0.6010 &
0.0177 &
0.0593 
&
0.9207$\rightarrow$0.9231 &
125.67 (0.219) \\

\scriptsize{\texttt{CF-VLA} (w/ r., round2)} &
0.6813 (\textbf{1.3898}) &
1.8291 (\textbf{3.7474}) &
0.6168 &
\textbf{0.0174} &
\textbf{0.0585} 
&
\textbf{0.9238}$\rightarrow$\textbf{0.9276} &
109.36 (0.123) \\

\bottomrule
\end{tabular}
} %
\vspace{-10pt}
\end{table*}

\subsection{Experimental Setup}
\noindent \textbf{Datasets.} 
We train and evaluate models on a large proprietary dataset consisting of 80,000 hours of human driving data from 25 countries, covering a variety of scenarios including highway and urban driving, weather conditions, and daytime and nighttime.
The entire data corpus forms the trajectory-only dataset $\mathcal{D}_\text{traj}$, which contains raw sensor data paired with ego-vehicle future trajectories.
Within $\mathcal{D}_\text{traj}$, we auto-labeled 3,000 hours of data and constructed the meta-action-labeled subset $\mathcal{D}_\text{meta}$.
Meta-action annotations are automatically extracted from expert trajectories through rule-based detectors operating on kinematic profiles. We label meta-actions at 10Hz over the balanced dataset curated based on Operational Design Domains (ODDs).
We split $\mathcal{D}_\text{meta}$ into training and validation subsets, with the validation subset denoted as $\mathcal{D}_\text{val}$.
All of our results are reported on this evaluation set.
The counterfactual reasoning dataset $\mathcal{D}_\text{CF}$ comes from the training set of $\mathcal{D}_\text{meta}$.
In summary, the base trajectory-only set $\mathcal{D}_\text{traj}$ contains about 11.6M 20-second video clips
and provides large-scale behavioral diversity.
For $\mathcal{D}_\text{meta}$, multiple samples are used for one video clip.
The training set of $\mathcal{D}_\text{meta}$ includes 433K 20s clips and 801K 8.4s samples. There are 39K video clips and 73K samples in the validation set.
The counterfactual reasoning dataset $\mathcal{D}_\text{CF}$ usually contains 200K samples.

\noindent \textbf{Metrics.}
We evaluate models along three dimensions: 

\noindent 1) \textbf{Trajectory Accuracy}:
We report \textit{MinADE/AvgADE} and \textit{MinFDE/AvgFDE} as mean/endpoint displacement errors over 6 predicted modes (lower is better), and \textit{Corner Distance} as the average deviation of vehicle-corner keypoints for measuring turning and lane-keeping precision.

\noindent 2) \textbf{Safety Characteristics}:
\textit{Collision Rate} measures the proportion of predicted trajectories that collide with other road users' trajectories within 5s,
while \textit{Out-of-road Rate} quantifies whether the predicted trajectories violate the road boundary. These complement distance-based metrics by revealing whether small deviations lead to unsafe outcomes.

\noindent 3) \textbf{Reasoning Quality}:
\textit{Meta-Action IOU} measures the alignment between predicted and ground-truth meta-actions over $64\times3$ bins (longitudinal, lateral, and lane). For CF-VLA, we report IOU after self-reflection, i.e., for the updated meta-actions. We also record \textit{Output Length} (\# tokens) and \textit{Think Rate}, the fraction of responses containing counterfactual reasoning, to quantify the test-time compute and adaptive reasoning.

\noindent

\noindent\textbf{Baselines.}
All models are initialized from \texttt{traj-only} for fair comparison. 
We prepare two variants of the models to train with or without route information, which contain 20 waypoints spanning the future 80m with equal spacing.
\begin{itemize}[leftmargin=*, itemsep=0pt, topsep=0pt]
    \item \texttt{traj-only}: trained purely on $\mathcal{D}_\text{traj}$ without any meta-action or reasoning signals, serving as the standard end-to-end Vision-Action model. No route is used.
    \item \texttt{meta-act}: introduces meta-action sequence as intermediate control primitives before trajectory generation.
    \item \texttt{lang-meta-act}: jointly predicts language reasoning, meta-actions, and trajectory. 
    \item \texttt{CF-VLA}: our fine-tuned models from \texttt{meta-act} and can perform counterfactual reasoning. The second round training of \texttt{CF-VLA} uses data from calling the data pipeline in \cref{sec:counterfactual} on the first round \texttt{CF-VLA}.
\end{itemize}

\begin{table*}[!t]
\centering
\caption{
\textbf{Ablations on meta–trajectory alignment and adaptive counterfactual reasoning.}
We train models without route information.
}
\label{tab:ablation-force-thinking}
\vspace{-3mm}
\small
\begin{tabular}{lcccccccc}
\toprule
\textbf{Model} &
\textbf{MinADE$\downarrow$} &
\textbf{AvgADE$\downarrow$} &
\textbf{MinFDE$\downarrow$} &
\textbf{AvgFDE$\downarrow$} &
\multicolumn{1}{c}{
\begin{tabular}[c]{@{}c@{}}\textbf{MinIOU$\uparrow$} \\(init $\rightarrow$ edited)\end{tabular}
}
&
\begin{tabular}[c]{@{}c@{}}\textbf{Corner}\\\textbf{Dist.$\downarrow$}\end{tabular} &
\begin{tabular}[c]{@{}c@{}}\textbf{Output}\\\textbf{Length}\end{tabular} &
\begin{tabular}[c]{@{}c@{}}\textbf{Think}\\\textbf{Rate}\end{tabular} \\
\midrule
\footnotesize{\texttt{meta-act} (baseline)} &
0.8411 &
1.6216 &
2.3647 & 
4.6616 &
0.9169  &
0.7720 &
85.32 &
-
\\
\footnotesize{\texttt{meta-act} (pre-filled)}	& 0.4831	& 0.9968	& 1.2412	& 2.5667 & 	1.0	& 0.4399 & 	8.00 & 	- \\
\midrule
\footnotesize{\texttt{CF-VLA} (adaptive)}    & \textbf{0.7650} & {1.5606} & \textbf{2.1416} & {4.3307} & 
0.9153$\rightarrow$\textbf{0.9212} & \textbf{0.6975} & {113.36} & 0.1478 \\
\footnotesize{\texttt{CF-VLA} (force no think)}	& 0.7897	& 1.4890	& 2.2178 & 4.1508 &	0.9133 	& 0.7274	&  87.43 &	0.0 \\
\footnotesize{\texttt{CF-VLA} (force think)}	& 0.9319	& 2.1144	& 2.8822	& 6.3699	& 
0.9132$\rightarrow$0.8565	& 0.8271 &	257.42	& 1.0 \\
\midrule
\footnotesize{\texttt{explicit} (meta$\rightarrow$act)}      & 0.7968 & \textbf{1.4686} & 2.2426 & \textbf{4.0922} & 0.9127 & 0.7363 & 87.99  & - \\
\footnotesize{\texttt{explicit} (CF reasoning)} & 0.9331 & 2.0628 & 2.8647 & 6.1872 & 0.8902$\rightarrow$0.8551 & 0.8339 & 258.06 & 0.9971 \\
\midrule
\footnotesize{\texttt{meta-act} (multi-round)}	& 0.7906	& 1.504	& 2.216	& 4.1995& 	0.9128	& 0.7275& 	88.3093 & 	- \\
\bottomrule
\end{tabular}
\end{table*}

\begin{table*}[!t]
\centering
\caption{
\textbf{Effect of our proposed data filtering pipeline.}
Models are fine-tuned with route information from \texttt{meta-act} (w/ route).
}
\label{tab:main-body-ablation-filter-data}
\vspace{-3mm}
\small
\begin{tabular}{lcccccccc}
\toprule
\textbf{Model} &
\textbf{MinADE$\downarrow$} &
\textbf{AvgADE$\downarrow$} &
\textbf{MinFDE$\downarrow$} &
\textbf{AvgFDE$\downarrow$} &
\multicolumn{1}{c}{
\begin{tabular}[c]{@{}c@{}}\textbf{MinIOU$\uparrow$} \\(init $\rightarrow$ edited)\end{tabular}
}
&
\begin{tabular}[c]{@{}c@{}}\textbf{Corner}\\\textbf{Dist.$\downarrow$}\end{tabular} &
\begin{tabular}[c]{@{}c@{}}\textbf{Output}\\\textbf{Length}\end{tabular} &
\begin{tabular}[c]{@{}c@{}}\textbf{Think}\\\textbf{Rate}\end{tabular} \\
\midrule

\footnotesize{\texttt{CF-VLA} (filtered ds)} &
{0.6712} &
1.4574 &
{1.7988} & 3.9466 &
0.9207$\rightarrow$0.9231 &
0.6010 &
125.67
 &
 0.2190
\\

\footnotesize{\texttt{CF-VLA} (whole ds)} &
{0.6811} &
1.4185 &
{1.8296} & 3.8344 &
0.9207$\rightarrow$0.9231 &
0.6128 &
191.14
 &
 0.6677
\\

\bottomrule
\end{tabular}
\vspace{-10pt}
\end{table*}

\subsection{Main Experiments}
\label{sec:main-exp}

We evaluate whether counterfactual reasoning improves trajectory accuracy, safety characteristics, and reasoning quality. Quantitative results are reported in \Cref{tab:openloop_results}.

\noindent\textbf{Impact of meta-actions and language.}
Starting from \texttt{traj-only}, introducing meta-actions (\texttt{meta-act} w/o route) reduces trajectory error by about 9\% in minADE and minFDE, showing that structured action abstraction provides stronger motion priors than direct trajectory generation.
Adding language supervision (\texttt{lang-meta-act}) yields a further $\sim$5\% improvement over \texttt{meta-act}, which indicates that language helps align motion intent with scene semantics.
With route information, \texttt{meta-act} (w/ route) provides an even stronger baseline.

\noindent\textbf{Impact of counterfactual reasoning.}
Compared to their non-reasoning counterparts, \texttt{CF-VLA} variants consistently improve both trajectory error and meta-action alignment.
Without route, \texttt{CF-VLA} (round2) achieves around 9-10\% lower minADE / minFDE than \texttt{meta-act} while also increasing meta-action IOU by roughly 0.5–1.0 absolute points after counterfactual editing.
With route, \texttt{CF-VLA} (round1) similarly improves over \texttt{meta-act} across trajectory, safety, and IOU, confirming that self-reflective reasoning leads to more expert-like meta-actions and trajectories.

\noindent\textbf{Behavioral safety.}
Gains from \texttt{CF-VLA} are not only geometric but also safety-critical.
Relative to \texttt{traj-only}, the best CF models reduce collision rate by roughly 25–30\% and off-road violations by about 15–20\%, while also lowering corner distance by $\sim$30\%.
Within each setting (with / without route), \texttt{CF-VLA} variants consistently achieve the lowest or near-lowest collision and off-road rates, indicating that counterfactual self-reflection translates into smoother, more stable, and more rule-consistent driving behavior.

\noindent\textbf{Impact of multi-round counterfactual training.}
Applying the data pipeline in \cref{sec:rollout-filter-label} a second time yields additional gains while making reasoning more efficient.
\texttt{CF-VLA} (w/o route, round2) outperforms \texttt{CF-VLA} (w/o route, round1) in AvgADE/FDE, off-road rate, and edited IOU, while using a lower think rate.
With route information, \texttt{CF-VLA} (w/ route, round2, 3 ds) slightly trades minimum error for better average ADE/FDE and higher IOU, and further improves collision and off-road rates compared to \texttt{CF-VLA} (w/ route, round1).
Importantly, the second-round models trained on 3 datasets reduce the think rate by almost half and shorten the average output length.
This shows that a second CF round can simultaneously refine performance and substantially reduce test-time overhead, extracting more value from the same driving data.

\noindent\textbf{Adaptive thinking and compute.}
Reasoning inevitably increases sequence length compared to non-reasoning models, but \texttt{CF-VLA} uses test-time compute much more efficiently than models that always think.
Compared with \texttt{lang-meta-act}, which reasons for every sample, \texttt{CF-VLA} (w/ route, round1) already achieves better performance with a think rate below 0.25.
After the second CF training round, \texttt{CF-VLA} (w/ route, round2, 3 datasets) further reduces the think rate by roughly 40–45\% while maintaining or improving average errors and IOU.
A similar pattern appears in the no-route setting, indicating that the model learns \textbf{{adaptive} reasoning}.
Another illustration is \cref{fig:teaser}.
It shows \textbf{\textit{{the model thinks more frequently when the scenario is harder}}}. The top plot shows that the think rate correlates strongly with the trajectory error (minADE). Simple scenarios like vehicle following rarely trigger counterfactual reasoning, while high-uncertainty or high-risk scenarios, such as lane changes, turning maneuvers, and VRUs, induce significantly more CF reasoning. 
CF-VLA achieves better error reduction when the scenarios are harder, suggesting that CF-VLA not only reasons adaptively but learns {when} reasoning is most beneficial.

\begin{tcolorbox}[
breakable,
  colback=black!2,
  colframe=black!20,
  arc=10pt,
  boxrule=0.8pt,
  left=5pt,
  right=5pt,
  top=3pt,
  bottom=0pt,
]
{\begin{center}\centering{\textbf{Key Takeaways}}\end{center}}
{
\vspace{-1.5em}
\begin{enumerate}[label=\textbf{\arabic*)}, leftmargin=8pt, itemsep=3pt, topsep=1pt]
    \item Self-reflection via counterfactual reasoning in \texttt{CF-VLA} consistently improves both the precision and reliability of driving planners, delivering the largest gains in accuracy, safety, and IOU beyond single-shot reasoning models.
  \item Multi-round CF training squeezes more value from the same data, improves performance while substantially reducing think rate, achieving the best accuracy–safety–compute trade-off.
  \item The results establish a clear performance ladder: 
    \\ \texttt{traj-only} $<$  \texttt{meta-traj}  \\
    $<$  \texttt{lang-meta-traj}  
    $<$ \texttt{CF-VLA}.
  \item CF-VLA learns adaptive reasoning: it saves computation on easy scenes and “thinks harder” in challenging scenarios, where counterfactual reasoning yields the largest error reductions.

\end{enumerate}
}
\end{tcolorbox}

\begin{figure*}
    \centering
    \includegraphics[width=\linewidth]{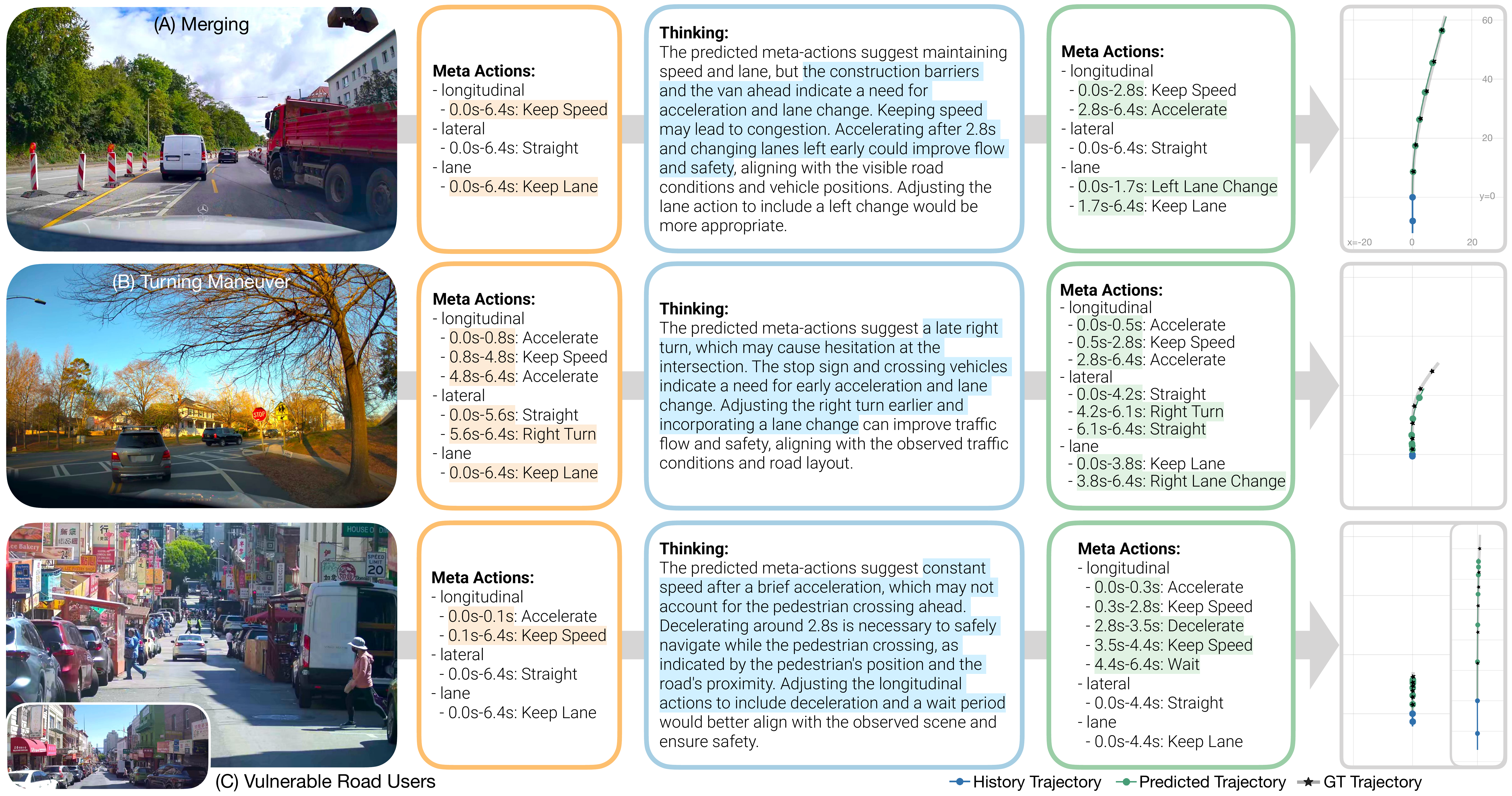}
    \vspace{-15pt}
    \caption{\textbf{Qualitative results of CF-VLA.} For three representative and safety-critical scenarios, each row shows the model's initial meta-actions \textcolor[RGB]{243,192,123}{(\textbf{left})}, the reasoning trace \textcolor[RGB]{173,204,224}{(\textbf{middle})}, and the updated meta-actions \textcolor[RGB]{166,204,170}{(\textbf{right})} together with the resulting trajectory. The counterfactual reasoning step identifies issues (missing lane changes, late turns, and failure to slow for pedestrians) and edits the meta-actions accordingly. 
    }
    \label{fig:qualitative}
    \vspace{-10pt}
\end{figure*}

\vspace{-5pt}
\subsection{Ablation Studies}
\label{sec:ablation-studies}

\noindent \textbf{Effect of meta-actions.}
\Cref{tab:openloop_results} has shown that adding meta-actions already improves over \texttt{traj-only}.
Here we focus on how well the \textbf{meta-trajectory alignment} is by comparing \texttt{meta-act} (baseline) and \texttt{meta-act} (pre-filled) in \Cref{tab:ablation-force-thinking}.
Pre-filling the meta-actions with ground truth nearly halves the trajectory error and substantially reduces corner distance.
This indicates that, once the meta-actions are correct, the model already has a strong meta-actions$\rightarrow$trajectory alignment, and most remaining error comes from imperfect meta-action prediction rather than the trajectory decoding.
This observation motivates conducting counterfactual reasoning directly on meta-actions.

\noindent \textbf{Effect of adaptive thinking.}
We study whether the adaptive reasoning control learned by the model improves reasoning quality and action accuracy.
We compare four variants:
(1) \texttt{CF-VLA} (force no think), where reasoning is disabled by forcing the model to emit ``Action:'' after meta-actions;
(2) \texttt{CF-VLA} (force think), where ``Thinking:'' is appended after meta-actions;
(3) \texttt{explicit} models that perform reasoning and action as two separate tasks under different user prompts; and
(4) the proposed \texttt{CF-VLA} (adaptive), which autonomously decides whether to self-reflect.
Results in \Cref{tab:ablation-force-thinking} show that always thinking increases computational cost and even degrades trajectory accuracy, while never thinking underperforms on difficult scenes.
The adaptive variant attains the best trade-off: lowest minADE among non-pre-filled models, higher edited IOU, and a moderate think rate, supporting that reasoning should be used selectively rather than uniformly, consistent with prior work on adaptive thinking~\cite{lin2025onetwovla}.

\noindent \textbf{Data filtering pipeline.}
We compare two \texttt{CF-VLA} models that differ only in this selection step.
In the \emph{Whole Dataset} variant, reasoning traces are generated for all meta-action–labeled samples.
In the \emph{Filtered Data} variant, CF traces are generated only for the subset that satisfies the trajectory-disagreement criterion (\cref{sec:rollout-filter-label}).
The \textit{Filtered Data} model achieves better minADE and minFDE and a lower corner distance, despite emitting much shorter responses and a lower think rate.
{These results indicate that counterfactual supervision must be targeted: simply adding more CF labels
and forcing the model to ``think'' on every scene introduces redundant or noisy reasoning signals,
diluting the impact of informative counterfactual examples and ultimately harming the performance.}
The rollout–filter–label stage is therefore not merely a data-efficiency optimization, but a crucial component for extracting reliable self-reflection signals.

\noindent \textbf{Multi-round training.}
We additionally train a model on the counterfactual filtered dataset, namely using
$(\mathcal{D}_\text{traj} \cup \mathcal{D}_\text{meta} \cup \mathcal{D}_\text{CF})$.
This model, \texttt{meta-act} (multi-round), is trained to predict only meta-actions and we discard the CF reasoning traces.
The resulting \texttt{meta-act} (multi-round) in \Cref{tab:ablation-force-thinking} shows that simply repeating high-value samples based on trajectory disagreement already yields a modest improvement over the single-round \texttt{meta-act} baseline.
However, \texttt{CF-VLA} brings larger gains in trajectory error and IOU, indicating that learning to edit meta-actions is more effective than only reusing them as fixed labels.

\subsection{Qualitative Results}
\vspace{-5pt}

\cref{fig:qualitative} visualizes the counterfactual self-reflection loop of CF-VLA in three representative scenarios. 
CF-VLA consistently identifies when its initial intent is misaligned with the scene and corrects it before trajectory generation. 
In (A) \emph{Merging}, the initial plan maintains speed and lane despite construction barriers and a slow van ahead, trapping the ego vehicle. The model opts for an early left lane change with acceleration to avoid congestion.
In (B) \emph{Turning Maneuver}, the model identifies the stop sign and crossing traffic and corrects a late right turn that causes hesitation at the intersection, yielding a more decisive and efficient maneuver.
In (C) \emph{Vulnerable Road Users}, the model notices a crossing pedestrian from the telephoto view and edits the dangerous actions to slow down and wait.
These cases show that CF-VLA’s self-reflection produces targeted, scene-grounded corrections that improve safety, traffic efficiency, and semantic consistency.

\section{Conclusion}
\vspace{-5pt}
We introduced Counterfactual VLA (CF-VLA), a self-reflective VLA framework that critiques and corrects its own actions before execution. 
A rollout–filter–label counterfactual pipeline allows CF-VLA to mine its own failure cases and improve over multiple training rounds. 
Experiments on large-scale driving datasets show consistent gains in trajectory accuracy, safety, and reasoning quality, demonstrating up to 17.6\% lower trajectory error and 20.5\% lower collisions than non-reasoning baselines.
The model demonstrates adaptive thinking: it reasons more in difficult, high-risk scenarios.
CF-VLA thus demonstrates that counterfactual self-reflection can effectively bridge reasoning and control, offering a general paradigm for autonomous driving systems that learn to think before they act.

{
    \small
    \bibliographystyle{ieeenat_fullname}
    \bibliography{main}
}

\clearpage
\setcounter{page}{1}
\maketitlesupplementary

This supplementary material is organized as follows.
\begin{itemize}
    \item 
\cref{appendix:ablation} presents additional ablation studies on dataset mixture, token-level loss weighting, and the effect of decoding temperature on adaptive counterfactual reasoning.
\item 
\cref{appendix:hyper} details the optimization setup, batching strategy, loss weighting, and sequence configuration used for all CF-VLA variants.
\item \cref{appendix:prompt} presents the full instruction prompts given to the VLA model during training and inference and to the VLM expert labeller for generating counterfactual reasoning traces.
\item 
\cref{appendix:visualization} provides extended qualitative visualizations, including both representative success cases and failure modes of CF-VLA’s counterfactual reasoning.
\end{itemize}

\section{Additional Ablation Studies}
\label{appendix:ablation}

\begin{figure}[H]
    \centering
    \includegraphics[width=0.8\linewidth]{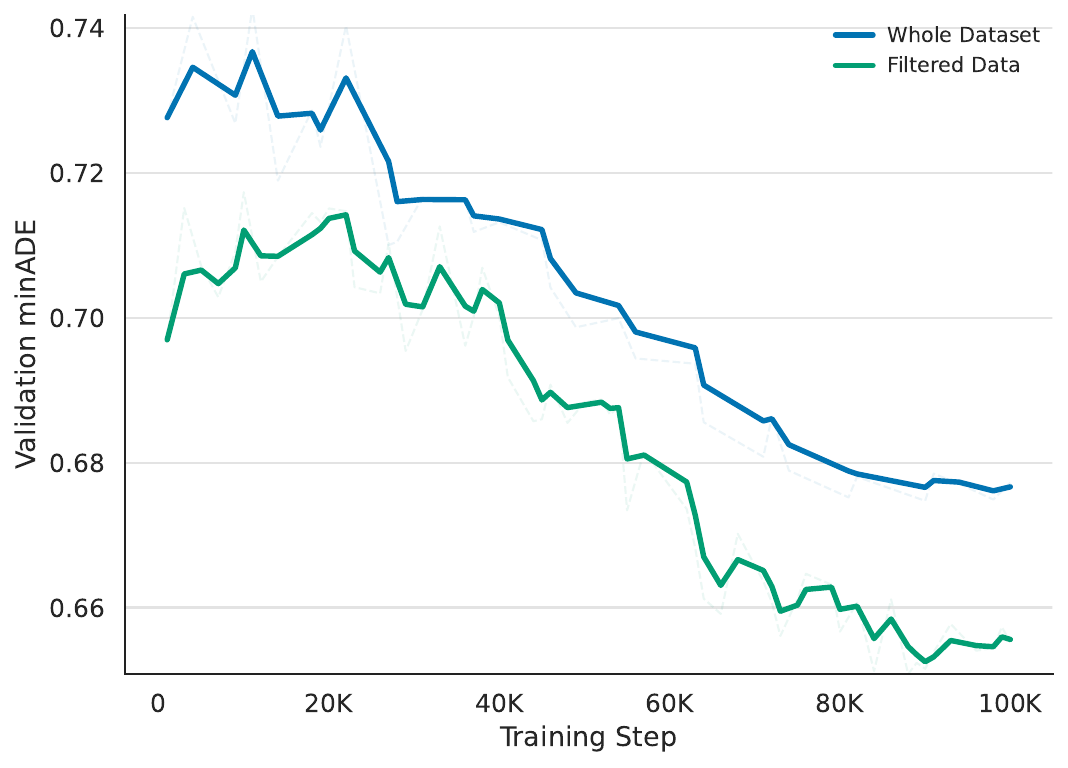}
    \caption{\textbf{Effect of counterfactual data filtering.}
    Validation minADE of CF-VLA on $\mathcal{D}_\text{val}$ when counterfactual (CF) labels are generated
    on the entire meta-action dataset (\textbf{Whole Dataset}) versus only on the high-value subset selected
    by the rollout–filter–label pipeline (\textbf{Filtered Data}).
    Training on CF labels for all scenes slows convergence and converges to a noticeably worse validation
    error, while restricting CF supervision to the filtered subset yields lower minADE throughout training
    and a better final model.}
    \label{appendix:fig:cf_filtering}
\end{figure}

\paragraph{Importance of counterfactual data filtering.}
The counterfactual pipeline deliberately applies CF supervision only to scenes where trajectory quality is bottlenecked by meta-actions, identified by the disagreement between free-generation and pre-filled meta-action rollouts.
A straightforward question is \textit{\textbf{what if we train the model to generate counterfactual reasoning for all data?}}
\cref{appendix:fig:cf_filtering} compares two \texttt{CF-VLA} models that differ only in this selection
step, and \cref{tab:ablation-filter-data} reports their final metrics.
In the \emph{Whole Dataset} variant, counterfactual reasoning traces are generated for \emph{all}
meta-action–labeled samples and used as CF data.
In the \emph{Filtered Data} variant, CF traces are generated only for the subset that satisfies the
trajectory-disagreement criterion (\cref{sec:rollout-filter-label}), i.e., cases where improved meta-actions are expected to
yield significant trajectory gains.
Despite using strictly more labeled CF data, the Whole Dataset model converges more slowly and
plateaus at a higher validation minADE.
The filtered model attains lower error across almost the entire training horizon and achieves the best
final minADE.

When counterfactual reasoning traces are generated only on the filtered subset, the model achieves
better minADE and minFDE and a lower corner distance, despite emitting much shorter responses and a lower think rate
(average length 125.7 vs.\ 191.1 tokens; think rate 0.22 vs.\ 0.67).
Generating CF traces on the entire dataset leads to substantially more and longer ``Thinking:'' segments
but does not improve, and in fact slightly degrades, key planning metrics.
\ul{These results indicate that counterfactual supervision must be targeted: simply adding more CF labels
and forcing the model to ``think'' on every scene introduces redundant or noisy reasoning signals,
diluting the impact of informative counterfactual examples and ultimately harming the performance.}
The rollout–filter–label stage is therefore not merely a data-efficiency optimization, but a crucial
component for extracting reliable self-reflection signals.

\begin{table*}[!t]
\centering
\caption{
\textbf{Effect of our proposed data filtering pipeline.}
We use models with route information and train them with the counterfactual reasoning traces on the filtered or the whole training dataset.
}
\label{tab:ablation-filter-data}
\vspace{-2mm}
\small
\begin{tabular}{lcccccccc}
\toprule
\textbf{Model} &
\textbf{MinADE$\downarrow$} &
\textbf{AvgADE$\downarrow$} &
\textbf{MinFDE$\downarrow$} &
\textbf{AvgFDE$\downarrow$} &
\multicolumn{1}{c}{
\begin{tabular}[c]{@{}c@{}}\textbf{MinIOU$\uparrow$} \\(init $\rightarrow$ edited)\end{tabular}
}
&
\begin{tabular}[c]{@{}c@{}}\textbf{Corner}\\\textbf{Dist.$\downarrow$}\end{tabular} &
\begin{tabular}[c]{@{}c@{}}\textbf{Output}\\\textbf{Length}\end{tabular} &
\begin{tabular}[c]{@{}c@{}}\textbf{Think}\\\textbf{Rate}\end{tabular} \\
\midrule

\footnotesize{\texttt{CF-VLA} (w/ route, filtered ds)} &
{0.6712} &
1.4574 &
{1.7988} & 3.9466 &
0.9207$\rightarrow$0.9231 &
0.6010 &
125.67
 &
 0.2190
\\

\footnotesize{\texttt{CF-VLA} (w/ route, whole ds)} &
{0.6811} &
1.4185 &
{1.8296} & 3.8344 &
0.9207$\rightarrow$0.9231 &
0.6128 &
191.14
 &
 0.6677
\\

\bottomrule
\end{tabular}
\end{table*}

\begin{figure*}[!t]
    \centering
    \includegraphics[width=\linewidth]{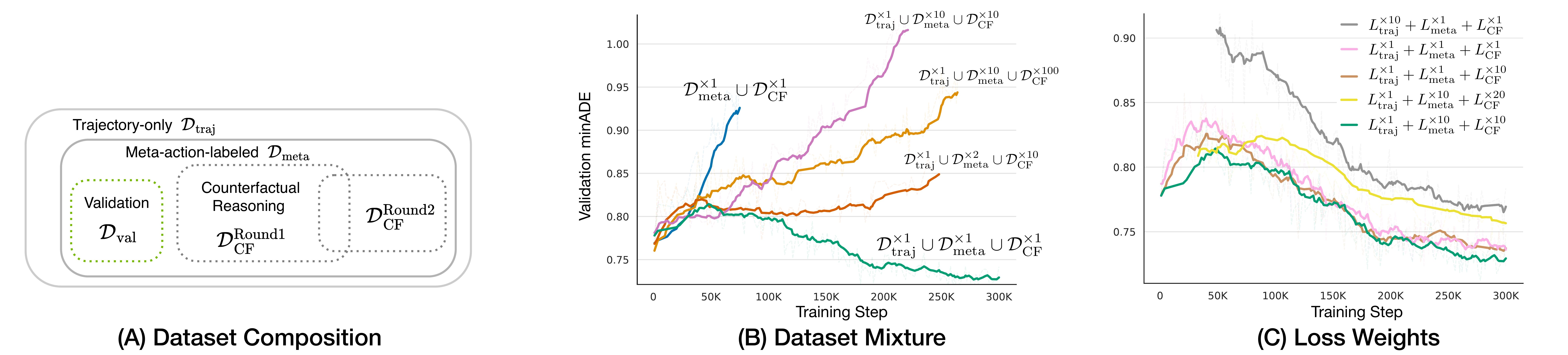}
    \caption{
    \textbf{(A)}\textbf{ Dataset composition.} A subset of the meta-action-labeled dataset $\mathcal{D}_\text{meta}$ is held out as the validation set $\mathcal{D}_\text{val}$.
    \textbf{(B)}\textbf{ Ablation on dataset mixture.} $\mathcal{D}^{\times x}$ denotes repeating the dataset $x$ times. For simplicity, $\mathcal{D}_\text{meta}$ denotes the training subset of meta-action-labeled data. Training with the natural combination of three tasks (\texttt{traj-only}, \texttt{meta-traj}, and \texttt{CF}) avoids overfitting and yields the best validation performance.
    \textbf{(C)}\textbf{ Ablation on loss weights.} $L^{\times y}$ applies a multiplier $y$ to the cross-entropy loss of the corresponding token group. Emphasizing meta-action and CF reasoning tokens with 
    $L_\text{traj}^{\times 1} + L_\text{meta}^{\times 10} + L_\text{CF}^{\times 10}$ stabilizes training and improves trajectory accuracy, whereas putting a large weight only on trajectory tokens degrades performance (the gray line). Further increasing the CF weight to
    $L_\text{traj}^{\times 1} + L_\text{meta}^{\times 10} + L_\text{CF}^{\times 20}$ induces more thinking (think rate $0.2338$ vs.\ $0.1478$ for the $1\!:\!10\!:\!10$ model) but harms trajectory accuracy, indicating that a higher think rate alone is not a reliable proxy for better control performance.
    }
    \label{appendix:fig:data_exp}
\end{figure*}

\paragraph{The impact of data mixture.}
As shown in \cref{appendix:fig:data_exp}(B), removing the large trajectory-only dataset $\mathcal{D}_\text{traj}$ and training only on meta-action and CF data (\textcolor[RGB]{48,113,173}{blue curve, $\mathcal{D}^{\times 1}_\text{meta} \cup \mathcal{D}^{\times 1}_\text{CF}$}) causes the model to overfit rapidly and generalize poorly.
Even when $\mathcal{D}_\text{traj}$ is included, aggressively repeating the smaller meta-action or CF datasets (\eg\ \textcolor[RGB]{197,101,38}{$\mathcal{D}^{\times 1}_\text{traj} \cup \mathcal{D}^{\times 2}_\text{meta} \cup \mathcal{D}^{\times 10}_\text{CF}$}, \textcolor[RGB]{203,133,189}{$\mathcal{D}^{\times 1}_\text{traj} \cup \mathcal{D}^{\times 10}_\text{meta} \cup \mathcal{D}^{\times 10}_\text{CF}$}, and \textcolor[RGB]{210,146,51}{$\mathcal{D}^{\times 1}_\text{traj} \cup \mathcal{D}^{\times 10}_\text{meta} \cup \mathcal{D}^{\times 100}_\text{CF}$}) eventually hurts validation minADE: the model degrades as duplicated meta-actions / CF traces dominate the training distribution.
The best configuration is the natural mixture of each dataset (\textcolor[RGB]{71,156,118}{green, $\mathcal{D}^{\times 1}_\text{traj} \cup \mathcal{D}^{\times 1}_\text{meta} \cup \mathcal{D}^{\times 1}_\text{CF}$}), where the data is sampled from the concatenated pool of all samples. This yields both the lowest validation error and the most stable convergence.
\ul{These results indicate that the large, diverse base trajectory dataset $\mathcal{D}_\text{traj}$ is essential for grounding the planner in realistic driving statistics, while a moderate amount of meta-action and counterfactual data is sufficient to teach the model how to verbalize and revise its plans; oversampling these specialized datasets reduces effective scene diversity and leads to worse trajectory accuracy due to overfitting.}

\paragraph{The impact of loss mixture.}
Token-level loss weighting is used to preserve the desired balance across sections (meta-actions, reasoning, and trajectory) in the model's response.
Denote the losses on trajectory, meta-action, and counterfactual reasoning tokens as $L_\text{traj}$, $L_\text{meta}$, and $L_\text{CF}$.
A configuration such as
$L_\text{traj}^{\times 1} + L_\text{meta}^{\times 10} + L_\text{CF}^{\times 10}$ (\cref{appendix:fig:data_exp}(C)) applies a $10\times$ multiplier to the meta-action and CF reasoning tokens relative to trajectory tokens and corresponds to the best-performing model.
As shown in \cref{appendix:fig:data_exp}(C), increasing the weights on meta-action and CF tokens ($L_\text{meta}^{\times 10}, L_\text{CF}^{\times 10}$) stabilizes training and improves reasoning consistency, whereas emphasizing trajectory tokens alone ($L_\text{traj}^{\times 10}$) deteriorates validation error.
However, pushing the CF weight further to
$L_\text{traj}^{\times 1} + L_\text{meta}^{\times 10} + L_\text{CF}^{\times 20}$ induces more thinking (raising the think rate from $0.1478$ to $0.2338$) but slightly harms trajectory accuracy.
\ul{This shows that think rate is not monotonically correlated with performance: overly rewarding CF tokens encourages longer and more frequent reasoning traces even when they do not translate into better meta-actions or trajectories.}
Balanced supervision between reasoning and control, rather than simply “thinking more”, is the key to strong performance.

\begin{figure}[H]
    \centering
    \includegraphics[width=\linewidth]{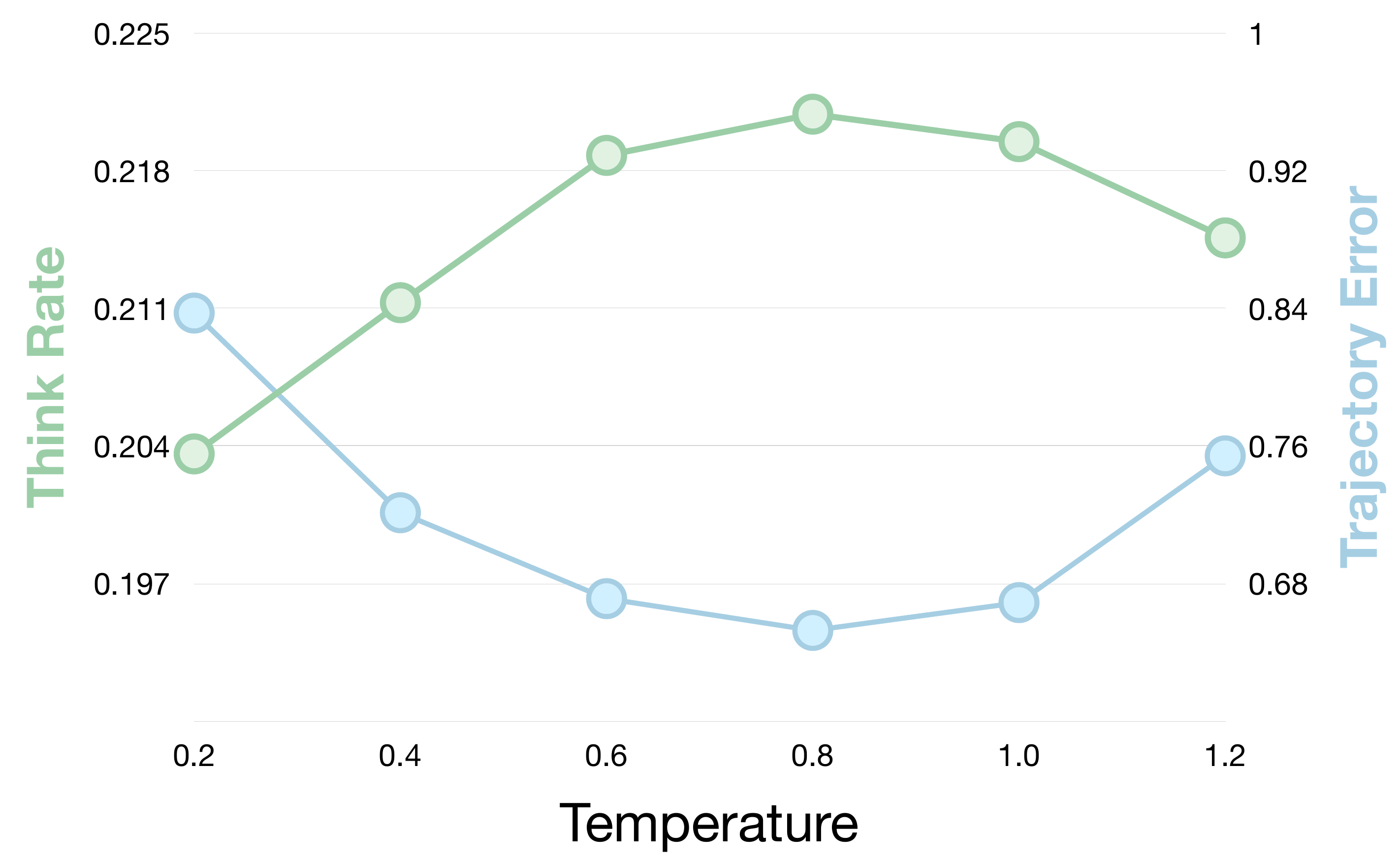}
\caption{\textbf{Effect of decoding temperature on adaptive counterfactual reasoning.}
Think rate \textcolor[RGB]{166,204,170}{(\textbf{left axis})} and trajectory error \textcolor[RGB]{173,204,224}{(\textbf{minADE, right axis})} of CF-VLA on the validation set under different sampling temperatures are plotted.
Both curves exhibit a strong inverse correlation: temperatures that induce higher think rates generally correspond to lower trajectory error, up to a moderate range around 0.8, while very low or very high temperatures either under-utilize counterfactual reasoning or introduce noisy generations that harm planning accuracy.}
\label{appendix:fig:temp_ablation}
\end{figure}

\begin{table*}[!t]
\centering
\caption{
\textbf{Ablation study on the training datasets.} 
We train \texttt{CF-VLA} (w/ route, round 2, 3 ds) with 3 datasets: $\mathcal{D}_\text{traj}$,$ \mathcal{D}_\text{meta}$,$\mathcal{D}_\text{CF}^\text{Round2}$ (3 ds).
We train \texttt{CF-VLA} (w/ route, round 2, 4 ds) with 
4 datasets: $\mathcal{D}_\text{traj}$,$ \mathcal{D}_\text{meta}$,$\mathcal{D}_\text{CF}^{\text{Round1}}$,$\mathcal{D}_\text{CF}^\text{Round2}$ (4 ds).
$\downarrow$ lower is better, $\uparrow$ higher is better.
}
\label{tab:ablation-34datasets}
\vspace{-2mm}
\small
{
\begin{tabular}{lccccccc}
\toprule
\textbf{Model} &
\begin{tabular}[c]{@{}c@{}}\textbf{ADE$\downarrow$} \\Min (Avg)\end{tabular} 
&
\begin{tabular}[c]{@{}c@{}}\textbf{FDE$\downarrow$} \\Min (Avg)\end{tabular} 
&
\begin{tabular}[c]{@{}c@{}}\textbf{Corner}\\\textbf{Dist.$\downarrow$}\end{tabular} &
\textbf{Collision$\downarrow$} &
\textbf{Off-road$\downarrow$} &
\begin{tabular}[c]{@{}c@{}}\textbf{IOU$\uparrow$}  \\init$\rightarrow$edited\end{tabular} 
&
\begin{tabular}[c]{@{}c@{}}\textbf{Output Len.}\\\textbf{(Think Rate)}\end{tabular}
\\
\midrule

\scriptsize{\texttt{meta-act} (w/ route)} &
0.7263 (1.4612) &
1.9561 (3.9269) &
0.6600 &
0.0196 &
0.0619 
&
0.9236 &
87.20 (--) \\

\scriptsize{\texttt{CF-VLA} (w/ route, round1)} &
\textbf{0.6712} (1.4574) &
\textbf{1.7988} (3.9466) &
0.6010 &
0.0177 &
0.0593 
&
0.9207$\rightarrow$0.9231 &
125.67 (0.219) \\

\scriptsize{\texttt{CF-VLA} (w/ r., round2, 3 ds)} &
0.6813 (\textbf{1.3898}) &
1.8291 (\textbf{3.7474}) &
0.6168 &
\textbf{0.0174} &
\textbf{0.0585} 
&
\textbf{0.9238}$\rightarrow$\textbf{0.9276} &
109.36 (0.123) \\

\scriptsize{\texttt{CF-VLA} (w/ r., round2, 4 ds)}&
0.6776 (1.4405) &
1.8108 ({3.9017}) &
\textbf{0.6083} &
0.0176 &
0.0588 
&
0.9186$\rightarrow$0.9241 &
140.23 (0.299) \\

\bottomrule
\end{tabular}
} %
\vspace{-10pt}
\end{table*}

\paragraph{Effect of decoding temperature.}
To study how decoding stochasticity interacts with adaptive counterfactual reasoning, this ablation varies the sampling temperature at inference while keeping all other settings fixed.
As shown in \cref{appendix:fig:temp_ablation}, trajectory error monotonically decreases when increasing the temperature from 0.2 to 0.8, but starts to rise again at 1.0 and 1.2, forming a U-shaped curve.
The think rate changes more mildly and peaks around 0.8.
\ul{Across temperatures, think rate and trajectory error are strongly and inversely correlated: configurations that elicit more counterfactual reasoning tend to produce more accurate trajectories.}
Very low temperatures make the model overly deterministic—once it settles on emitting ``Action:'' for many scenes, it under-explores the ``Thinking:'' branch, limiting the benefit of self-reflection and leading to suboptimal trajectories.
In contrast, overly high temperatures inject excessive randomness into meta-actions and reasoning traces, which degrades trajectory quality despite a similar average think rate.
A moderate temperature therefore provides the best trade-off, preserving stable planning while allowing CF-VLA to invoke counterfactual reasoning where it is most beneficial.

\paragraph{Training dataset composition.}
As shown in \cref{tab:ablation-34datasets}, comparing \texttt{CF-VLA} (w/ route, round 2, 3 datasets) and \texttt{CF-VLA} (w/ route, round 2, 4 datasets) reveals an important efficiency trade-off.
Including the first-round CF data (4 datasets) slightly improves geometric metrics, but degrades average errors, IOU, and safety metrics, and nearly doubles the think rate and output length.
This suggests simply adding more CF traces (especially when they are not on-policy rolled out) is not always beneficial, and \texttt{CF-VLA} (w/ route, round2, 3 datasets) offers a better balance between accuracy, safety, and test-time efficiency.

\section{Training Hyper-parameters}
\label{appendix:hyper}

All CF-VLA variants are trained from the \texttt{Qwen2.5-VL-3B-Instruct} backbone using the optimization and data settings listed below.

\paragraph{Optimization.}
Models are optimized with AdamW with $\beta_1 = 0.9$, $\beta_2 = 0.95$, $\epsilon = 10^{-8}$ and weight decay $0.01$.
The initial learning rate is $1\times 10^{-5}$ with a cosine decay schedule and $2{,}000$ warm-up steps.
For \texttt{CF-VLA} models, we set the learning rate to $5\times 10^{-6}$.
Training runs for $300{,}000$ optimization steps with gradient norm clipping at $1.0$.

\paragraph{Batching and precision.}
The per-device training batch size is $1$ with gradient accumulation set to $1$ (global batch size equal to the number of GPUs).  
Training uses bfloat16 without gradient checkpointing and activation checkpointing.  
Distributed training uses fully sharded data parallelism. The VLM side uses FlashAttention-2 for all attention layers with bfloat16 activations. We use $64$ NVIDIA A100 GPUs and thus batch size is $64$ for each experiment.

\paragraph{Loss weighting and data mixture.}
Training minimizes token-level cross-entropy over the assistant outputs only. Different token groups are weighted as
\[
w_{\text{act}} : w_{\text{meta}} : w_{\text{CF}} = 1 : 10 : 10,
\]
for trajectory tokens (\texttt{act}), meta-action tokens (\texttt{meta}), and counterfactual reasoning tokens (\texttt{cot}), respectively.
For counterfactual samples, the loss on the first (uncorrected) meta-action block is masked out.  
Unless otherwise specified, the \texttt{meta-act} models are trained on the mixture $D_{\text{traj}} \cup D_{\text{meta}}$ and CF-VLA models are trained on the mixture $D_{\text{traj}} \cup D_{\text{meta}} \cup D_{\text{CF}}$ without oversampling (repetition) of any dataset.

\paragraph{Sequence configuration.}
Each training example uses $4$ frames per camera over a $2$s history window (2Hz), with resolution set to $448\times796$. The model inputs $16$ history waypoints and predicts $64$ future waypoints.

\clearpage
\onecolumn
\section{Instruction Template}
\label{appendix:prompt}

The following box shows the full instruction prompt (system and user messages) used to query the VLA models for producing meta-actions and the optional counterfactual reasoning traces during training and evaluation:
\begin{small}
\begin{promptbox}
(*@\textbf{<|im\_start|>system}@*)
You are a helpful autonomous driving assistant.

You will receive videos and a history trajectory as context and the goal is to produce future trajectories that are reasonable in the driving scene, safe and consistent with the history trajectory.

The videos are captured by cameras mounted on the self-driving car, showing the driving scene over the past 2 seconds. Camera camera_front_wide_120fov shows the front of the car with a 120 degree field of view. Camera camera_front_tele_30fov shows the front of the car with a 30 degree field of view, the zoom-in view of camera_front_wide_120fov.

The history trajectory is 1.6 seconds of past motion for the self-driving car, sampled at 10 Hz, and includes x, y, z coordinates and heading (all four values are relative to the state at the current time). It is enclosed between <|traj_history_start|> and <|traj_history_end|>, and is encoded into a continuous latent representation using 1 token by the history trajectory encoder.

The future trajectory covers 6.4 seconds of predicted motion, also sampled at 10 Hz, and includes x, y, z coordinates and heading (relative to the state at the current time). The future trajectory is represented by a set of 6 discrete tokens enclosed between <|traj_future_start|> and <|traj_future_end|>. Each token is sampled from the vocabulary of the future trajectory tokenizer, which will decode the 6 tokens into a 6.4-second future trajectory.

You may receive different types of tasks given the context, including:
- Producing a future trajectory given the video and history trajectory.
- Proposing the meta actions in the driving process and generate the future trajectory according to the meta actions.
- Reflecting on the correctness of the meta actions according to the information in the scene and correcting the meta actions if necessary.

Meta actions are a set of high-level descriptions of the vehicle's behavior. Meta actions are grouped into three categories:
- **longitudinal**: ["Accelerate", "Decelerate", "Keep Speed", "Wait", "Reverse"]
- **lateral**: ["Straight", "Left Turn", "Right Turn"]
- **lane**: ["Keep Lane", "Left Lane Change", "Right Lane Change"]

The meta actions must partition the 6.4-second planning horizon (from 0.0s to 6.4s) without gaps or overlaps *within each group*. Groups are optional; omit any group not relevant to the current scene. List the meta actions in each group in bullet-point format, like the example below:

- longitudinal
  - 0.0s-0.9s: Keep Speed
  - 0.9s-6.4s: Accelerate
- lateral
  - 0.0s-6.4s: Straight
- lane
  - 0.0s-4.4s: Left Lane Change
  - 4.4s-6.4s: Keep Lane

Each time interval (in seconds) is specified as `start-end`, followed by a colon and an action within the group.

If you are tasked to propose the meta actions, you should first generate the meta actions before generating the future trajectory. If you think the first proposed meta actions are **incorrect**, generate a "Thinking:" section that briefly reflects on your meta-actions and explains the problem. Then provide a second "Meta Actions:" section with a corrected meta-action plan, and finally generate the future trajectory in the "Action:" section. If you are confident that the first meta actions are **valid**, you may skip the reflection section and go directly to the trajectory generation.

Respond appropriately based on the given input and task.(*@\textbf{<|im\_end|>}@*)
(*@\textbf{<|im\_start|>user}@*)
The video for camera 1 (camera_front_wide_120fov) is: <|vision_start|><|video_pad|>...<|vision_end|>The video for camera 6 (camera_front_tele_30fov) is: <|vision_start|><|video_pad|>...<|vision_end|>The trajectory history is: <|traj_history_start|><|traj_history|><|traj_history_end|>First, list the meta actions in the driving process. Then, generate a possible future trajectory. If the meta-action is unsafe or incorrect, reflect on it and provide a corrected one. Use the output format below.

Meta Actions:
[meta actions]

Thinking: (optional)
[text]

Meta Actions: (optional)
[corrected meta actions]

Action:
<|traj_future_start|>[6 tokens]<|traj_future_end|>(*@\textbf{<|im\_end|>}@*)
(*@\textbf{<|im\_start|>assistant}@*)
\end{promptbox}
\end{small}

The following box shows the full instruction prompt (system and user messages) given to the VLM expert labeller (\texttt{Qwen2.5-VL-72B-Instruct}) for generating counterfactual reasoning traces and refining meta-action annotations:
\begin{small}
\begin{promptbox}
(*@\textbf{<|im\_start|>system}@*)
You are an autonomous-driving labeling assistant. Your job is to **diagnose and correct** prediction errors **using visual cues** in the provided frames/videos. You **must not use** or reveal any privileged knowledge beyond what is visually observable. Be concrete, timestamped, and concise.(*@\textbf{<|im\_end|>}@*)
(*@\textbf{<|im\_start|>user}@*)
You will be presented with a driving scenario where you need to choose the correct meta-actions. Your task is to analyze the situation and provide reasoning about why the previous meta-actions are less preferable to the expert meta-actions.

## Observation

You will receive: the PREDICTED meta-actions (longitudinal, lateral, lane) with time intervals as well as the EXPERT meta-actions. Meta-actions are a set of high-level descriptions of the vehicle's behavior. Meta-actions are grouped into three categories with these options in each category:

- **longitudinal**: ["Accelerate", "Decelerate", "Keep Speed", "Wait", "Reverse"]
- **lateral**: ["Straight", "Left Turn", "Right Turn"]
- **lane**: ["Keep Lane", "Left Lane Change", "Right Lane Change"]

The meta-actions must partition the 6.4-second planning horizon (from 0.0s to 6.4s). Groups are optional - omit any group not relevant to the current scene. The meta-actions are in bullet-point format. Each time interval (in seconds) is specified as `start-end`, followed by a colon and an action within the group. Like the example below (note that this example is not related to the scene you are watching):

- longitudinal
  - 0.0s-0.9s: Keep Speed
  - 0.9s-6.4s: Accelerate
- lateral
  - 0.0s-6.4s: Straight
- lane
  - 0.0s-4.4s: Left Lane Change
  - 4.4s-6.4s: Keep Lane

You will receive: VIDEOS/FRAMES from two cameras. The images are captured by cameras mounted on the self-driving car, showing the driving scene at this moment. Camera `camera_front_wide_120fov` shows the front of the car with a 120 degree field of view. Camera `camera_front_tele_30fov` shows the front of the car with a 30 degree field of view, the zoom-in view of `camera_front_wide_120fov`.

## Task

Your task is to provide a detailed counterfactual reasoning trace as an internal self-reflection that demonstrates your reasoning process for the current situation. Your reasoning should:

1. Start with analyzing the driving scenario and the goal. Highlight any relevant visual clues, constraints, or consequences from the scenario. For example, the lane markings, the traffic lights, the vehicles, the pedestrians, the road conditions, etc.
2. Discuss the predicted meta-actions, explain why they may be less optimal, think about the expected consequences. If the predicted meta-actions are already close to the expert meta-actions, it's OK to simply acknowledge that the meta-actions are already good and explain why.
3. Justify why the expert action may be more preferable, while not indicating you have access to the expert action. State the possible changes to the predicted meta-actions to make it closer to the expert action.


Hard rules:

- **Never mention** ground truth, labels, "expert", "GT", "dataset", or phrases implying access to future or privileged information. The ground truth meta-actions are only used to help you understand the scene and the predicted meta-actions. Do not discuss the ground truth meta-actions in the reasoning to avoid information leakage. The ultimate goal is to insert the reasoning after the predicted meta-actions and before the GT meta-actions so I can build the dataset that have 1) wrong meta-actions, 2) the reflection, and 3) the corrected meta-actions. Therefore, it's strictly forbidden to imply that you have access to the ground truth meta-actions in the reasoning. Do not discuss anything like "according to the images", "the trajectory suggests" or "the GT meta-actions suggest", "the vehicles seem like [some future actions]", etc.
- **Anchor** every claimed correction with one or more specific cues. Always ground your decisions based on the provided images. Discuss the reasoning in the context according to the given images. Do not simply say "which meta action is wrong and I should change it to something".
- Use **one short paragraph** for `reasoning` ( (*@$\leq$@*) 80 words, no line breaks). Your reasoning should be concise and to the point. Do not list the final (correct) meta-actions. Do not use markdown or other formatting in the reasoning. Avoid meta-commentary about being an AI. Use natural, step-by-step reasoning. Focus on logical decision-making. Directly write the self-reflection reasoning, no extra headings, disclaimers, or external notes.
- If uncertain or the predicted meta-actions are similar to the ground truth meta-actions, you can simplify the reasoning.
- It's OK to find that the predicted meta-actions are already close to the expert meta-actions, and you don't need to change too much.

## Scenario Data
The video captured by camera_front_wide_120fov: <|vision_start|><|video_pad|>...<|vision_end|>

The video captured by camera_front_tele_30fov: <|vision_start|><|video_pad|>...<|vision_end|>

The expert meta-actions are:
- longitudinal
  - 0.0s-0.5s: Keep Speed
  - 0.5s-5.0s: Decelerate
  - 5.0s-5.9s: Keep Speed
  - 5.9s-6.4s: Accelerate
- lateral
  - 0.0s-5.0s: Left Turn
  - 5.0s-6.4s: Straight
- lane
  - 0.0s-5.5s: Keep Lane
  - 5.5s-6.4s: Left Lane Change

The predicted meta actions are:
- longitudinal
  - 0.0s-6.4s: Keep Speed
- lateral
  - 0.0s-3.4s: Left Turn
  - 3.4s-6.4s: Straight
- lane
  - 0.0s-6.4s: Keep Lane

Please propose counterfactual reasoning on the predicted meta-actions.(*@\textbf{<|im\_end|>}@*)
(*@\textbf{<|im\_start|>assistant}@*)
\end{promptbox}
\end{small}

\clearpage
\onecolumn
\section{Additional Visualization}
\label{appendix:visualization}

In this section, we provide more visualization results for model \texttt{CF-VLA} (w/ route, round 2, 3 datasets).

\begin{figure}[H]
    \centering
    \includegraphics[width=\linewidth]{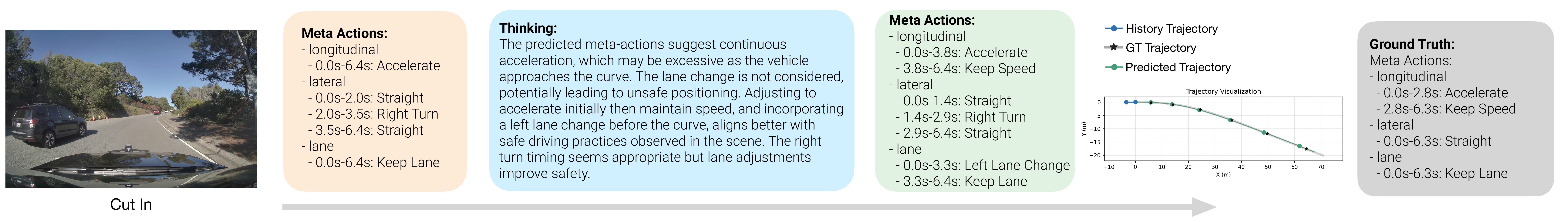}
    \vspace{-1.5em}
    \caption{
\textbf{Counterfactual reasoning for a cut-in near a curve.}
The scene shows the ego vehicle following traffic on a two-lane road where a right-hand curve and potential cut-in are ahead. 
The initial meta-actions (orange) prescribe continuous acceleration through the entire horizon and keep the ego in its current lane while executing the right turn, which could build up unnecessary speed as the vehicle enters the curve and leaves no lateral margin to surrounding traffic. 
Conditioned on these meta-actions and the video, the counterfactual reasoning step (blue) recommends accelerating only in the early phase and then maintaining speed, while also introducing a left lane change before the curve to obtain a safer positioning. 
The revised meta-actions (green) thus shorten the acceleration period, add a brief left lane change followed by lane keeping, and retain a similar right-turn timing; the resulting trajectory remains very close to the ground truth (gray) in space while adopting a more explicitly safety-oriented lane strategy. 
This example highlights how CF-VLA can simultaneously reshape longitudinal and lane-level intent around curves and cut-ins, yielding a conservative yet efficient plan that stays consistent with the scene.
}
\end{figure}

\begin{figure}[H]
    \centering
    \includegraphics[width=\linewidth]{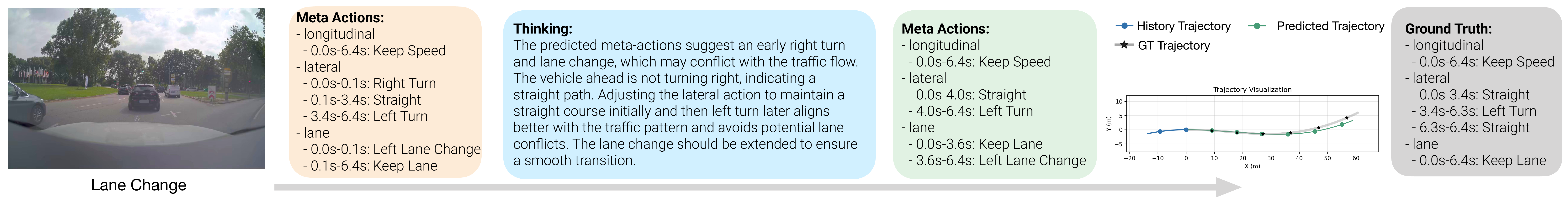}
    \vspace{-1.5em}
    \caption{
\textbf{Self-correction in a lane-change scenario.}
The scene shows the ego vehicle following a lead car on an urban road while approaching a junction where the traffic ahead continues straight and then bends left. 
The initial meta-actions (orange) incorrectly introduce a brief right turn and lane change at the beginning of the horizon (around $0$–$0.1\,$s), followed by a later left turn, creating an unnecessarily complex maneuver that could conflict with the surrounding traffic pattern. 
Conditioned on these meta-actions and the video frames, the counterfactual reasoning step (blue) notes that the lead vehicle is not turning right and recommends maintaining a straight lateral course for longer before initiating a smoother left turn and extended lane change. 
The revised meta-actions (green) thus remove the spurious early right turn, delay the left-turn phase to around $4.0\,$s, and stretch the lane-change interval, yielding a predicted trajectory that closely follows the ground-truth path (gray) while preserving similar speed. 
This example illustrates how CF-VLA can use visually grounded counterfactual reasoning to correct an initially misaligned turning intent and produce a more coherent, traffic-consistent plan.
}
\end{figure}

\begin{figure}[H]
    \centering
\includegraphics[width=\linewidth]{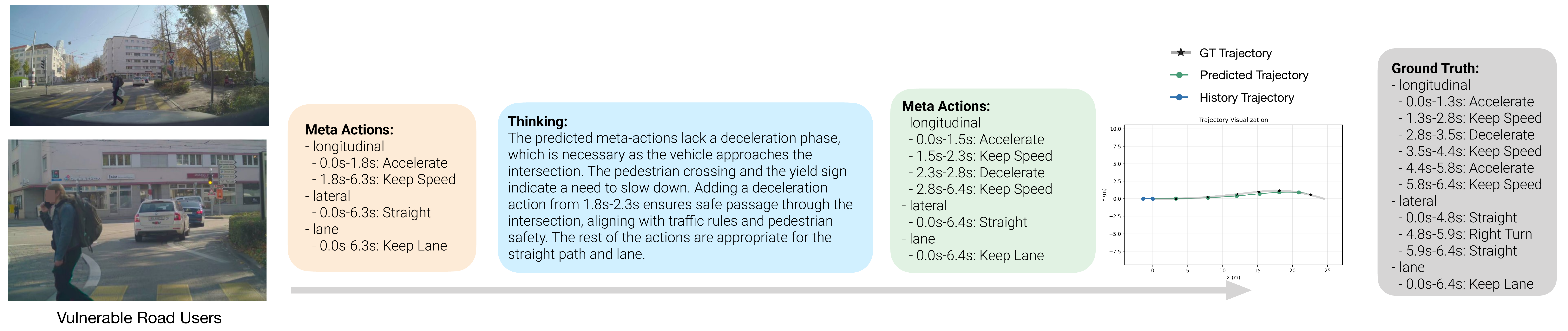}
    \vspace{-1.5em}
    \caption{
\textbf{Counterfactual reasoning near vulnerable road users and queuing traffic.}
The scene shows a pedestrian finishing a crossing at an intersection while a queue of vehicles waits ahead. 
The initial meta-actions (orange) instruct the ego to accelerate from standstill and then maintain speed from $1.8$–$6.3\,$s, lacking any explicit deceleration phase before reaching the intersection and the queue. 
Conditioned on these meta-actions and the telephoto views, the reasoning step (blue) notes the pedestrian crossing and yield signage and recommends inserting a short deceleration interval, yielding revised meta-actions (green) that accelerate from $0$–$1.5\,$s, briefly cruise, then decelerate between $2.3$–$2.8\,$s before keeping speed while joining the line of cars. 
The resulting trajectory closely follows the longitudinal profile, where the human driver also accelerates once the pedestrian has cleared the crosswalk and then slows to queue at the intersection.
This example illustrates that CF-VLA’s self-reflection can enrich an initially over-optimistic plan with an appropriate queuing deceleration, achieving behavior that is both pedestrian-aware and consistent with surrounding traffic.
}
\end{figure}

\begin{figure}[H]
    \centering
    \includegraphics[width=\linewidth]{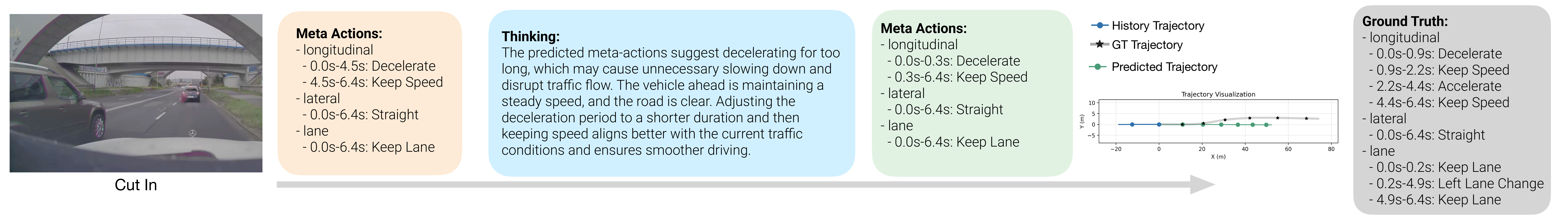}
    \caption{
\textbf{Conservative counterfactual reasoning in a cut-in scenario.}
The scene shows highway driving with a lead vehicle in the ego lane and another vehicle overtaking in the adjacent left lane. 
The initial meta-actions (orange) propose a long deceleration from $0$–$4.5\,$s followed by keeping speed, even though the lead vehicle is already maintaining a steady pace and the road ahead is largely clear. 
Conditioned on these meta-actions and the videos, the reasoning step (blue) correctly identifies this as overly cautious and recommends shortening the deceleration period, switching to a plan that briefly decelerates from $0$–$0.3\,$s and then keeps speed (green), yielding a smoother and less disruptive longitudinal profile that better matches the traffic flow. 
However, both the initial and revised meta-actions retain a ``Keep Lane'' decision for the full horizon, whereas the human driver (gray) performs a {left lane change} and subsequent acceleration to pass slower traffic. 
As a result, the predicted trajectory remains safe but more conservative and less efficient than the human expert, likely prioritizing collision avoidance with the adjacent overtaking vehicle over an aggressive lane change.
}
\end{figure}

\begin{figure}[H]
    \centering
    \includegraphics[width=\linewidth]{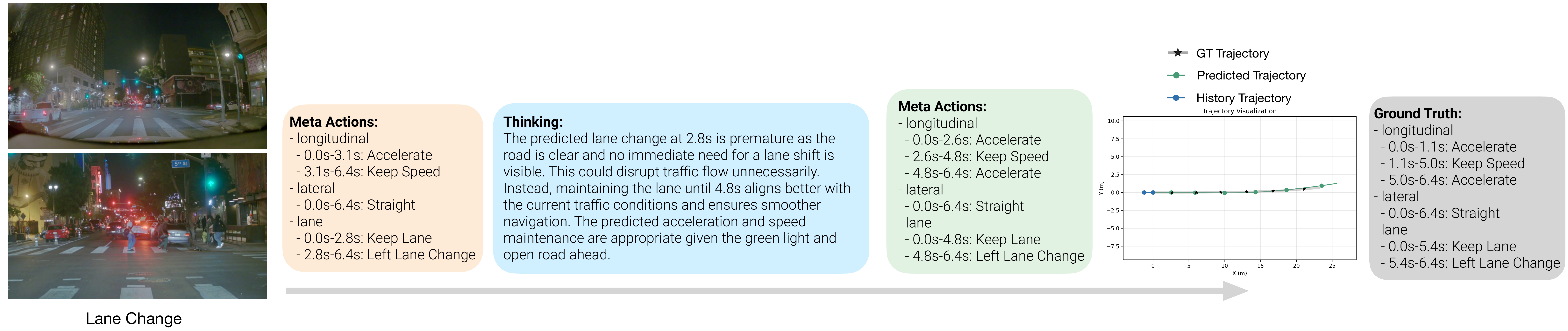}
    \caption{
\textbf{Refining lane-change timing with counterfactual reasoning at night.}
The scene shows nighttime urban driving through a signalized intersection with pedestrians near the crosswalk and open road beyond. 
The initial meta-actions (orange) plan to accelerate through the green light and initiate a left lane change at $2.8\,$s, causing the ego vehicle to start shifting lanes shortly after entering the intersection. 
Conditioned on these meta-actions and the video input, the counterfactual reasoning step (blue) judges this maneuver as premature given the clear lane ahead and potential disruption to traffic, and recommends extending the keep-lane phase before moving left. 
The revised meta-actions (green) therefore maintain lane until about $4.8\,$s, then perform a later left lane change while preserving a smooth accelerate–cruise–accelerate longitudinal pattern. 
As shown in the trajectory visualization, this produces a predicted trajectory whose lane-change timing closely matches the human driver (gray), who also stays in the original lane until well past the intersection before merging left. 
This example highlights how CF-VLA can use counterfactual self-reflection to delay unnecessary early lane changes, leading to behavior that is both smoother and more consistent with human driving.
}
\end{figure}

\begin{figure}[H]
    \centering
    \includegraphics[width=\linewidth]{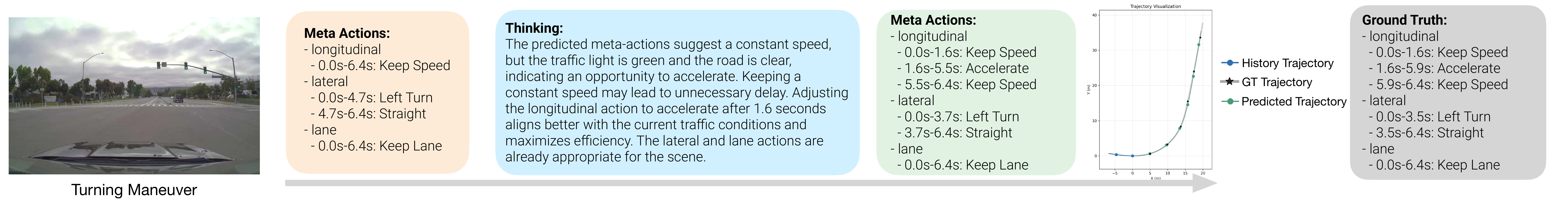}
    \caption{
\textbf{Refining acceleration timing in a turning maneuver with counterfactual reasoning.}
The scene shows the ego vehicle approaching a large signalized intersection with a green light and no visible cross-traffic. 
The initial meta-actions (orange) plan to keep a constant speed for the entire $6.4\,$s horizon while executing a left turn followed by a straight segment, which is safe but under-utilizes the open intersection and may lead to unnecessary delay. 
Conditioned on these meta-actions and the video input, the counterfactual reasoning step (blue) correctly notes that the road is clear under green and recommends introducing an acceleration phase after roughly $1.6\,$s, while leaving the lateral and lane decisions unchanged. 
The revised meta-actions (green) therefore maintain speed initially, then accelerate through the middle of the intersection before returning to a cruising speed, producing a predicted trajectory that closely matches the human driver’s motion (gray) in both lateral path and longitudinal profile. 
This example illustrates how CF-VLA can inject a well-timed mid-intersection acceleration through self-reflection, improving efficiency without sacrificing safety.
}
\end{figure}

\newpage
We also show the failure cases:

\begin{figure}[H]
    \centering
    \includegraphics[width=\linewidth]{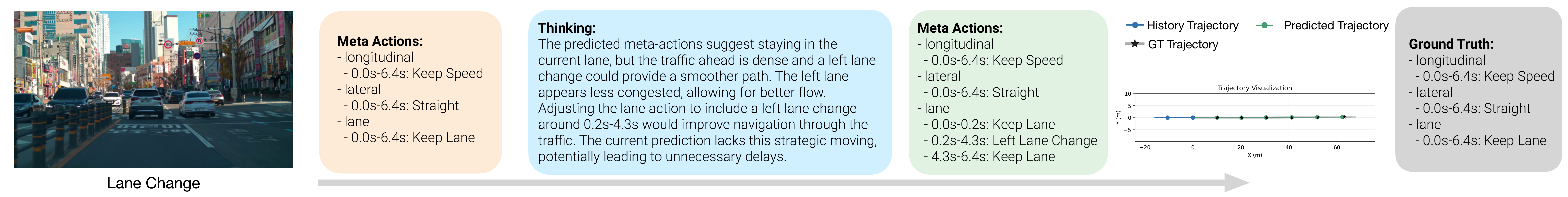}
    \caption{
\textbf{Failure case of CF-VLA's counterfactual reasoning at a straight lane.}
The scene shows the ego vehicle on a straight multi-lane urban road with moderate traffic ahead. 
The initial meta-actions (orange) propose a reasonable plan that keeps speed and lane throughout the horizon. 
However, conditioned on these meta-actions and the video input, the counterfactual reasoning step (blue) argues that the left lane ``appears less congested’’ and recommends inserting a left lane change between $0.2$\,s and $4.3$\,s to ``improve navigation through the traffic.’’ 
The revised meta-actions (green) follow this suggestion and introduce a left lane change, causing the predicted trajectory to drift into the adjacent lane, while the ground-truth driver in fact keeps lane and continues straight, as indicated by the trajectory visualization and ground-truth meta-actions (gray). 
This example illustrates that CF-VLA can occasionally over-correct a safe plan based on a misinterpretation of visual cues, hallucinating a beneficial lane change where the human driver does not perform one. 
Such failure cases highlight the need for better calibration between counterfactual reasoning, route-level intent, and true traffic semantics.
}
\end{figure}

\begin{figure}[H]
    \centering
    \includegraphics[width=\linewidth]{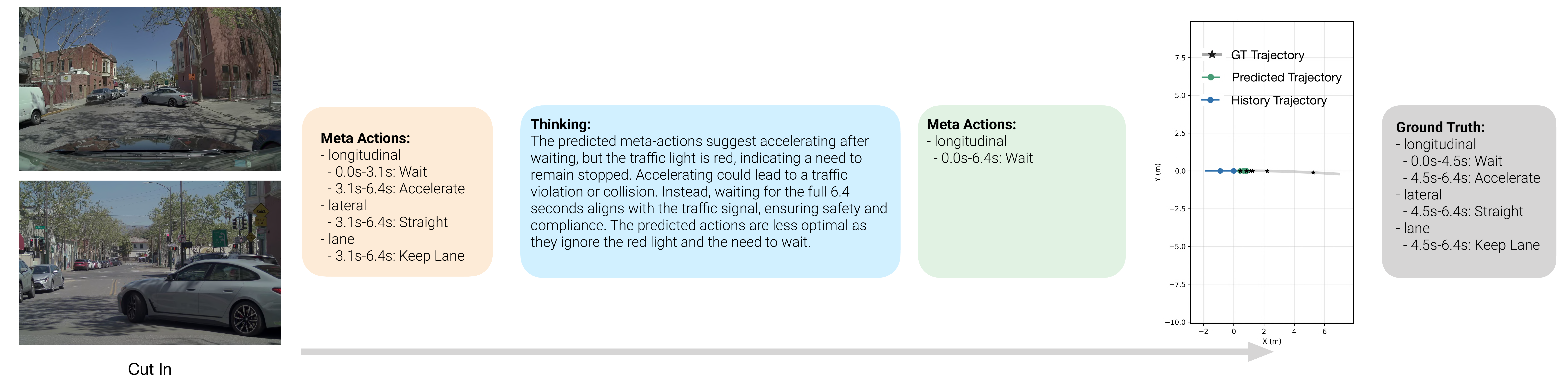}
\caption{
\textbf{Failure case of CF-VLA's counterfactual reasoning at a cut-in.}
The scene shows the ego vehicle waiting while a nearby vehicle cuts into the lane ahead at an urban intersection. 
The initial meta-actions (orange) propose to wait until about $3.1\,$s and then accelerate straight in the current lane from $3.1$–$6.4\,$s, which qualitatively resembles the human driver who eventually moves off when the vehicle above leaves. 
Conditioned on these meta-actions and the video frames, the counterfactual reasoning step (blue) incorrectly concludes that any acceleration would ``ignore the red light’’ and recommends waiting for the full $6.4$s to ensure safety and compliance. 
The revised meta-actions (green) therefore switch to a pure ``wait'' plan over the entire horizon, producing a predicted trajectory that remains stopped, whereas the ground-truth trajectory (gray) waits only until about $4.5\,$s before accelerating and clearing the intersection. 
The model misinterprets the scenario by observing the traffic light and ignores that the major factor causing the waiting behavior is the vehicle above.
This example illustrates how CF-VLA's self-reflection can sometimes become overly conservative and miscalibrated with respect to the spatial relationship between objects and traffic lights, degrading performance by overriding a reasonable initial plan.
}
\end{figure}

\end{document}